\documentclass{article}

 \pdfoutput=1

\usepackage[preprint]{neurips_2023}

\usepackage{amsmath}
\usepackage[utf8]{inputenc} 
\usepackage[T1]{fontenc}    
\usepackage{hyperref}       
\usepackage{url}            
\usepackage{booktabs}       
\usepackage{amsfonts}       
\usepackage{nicefrac}       
\usepackage{microtype}      
\usepackage{xcolor}         
\usepackage[pdftex]{graphicx}

\usepackage{mathtools}

\usepackage{tikz}
\usetikzlibrary{tikzmark,decorations.pathreplacing,calligraphy,positioning,calc,arrows.meta,shapes.geometric,shapes.arrows}

\title{Demographic Parity: \\Mitigating Biases in Real-World Data}

\author{%
  Orestis Loukas\\
  \texttt{orestis.loukas@staff.uni-marburg.de} \\
  \And
  Ho Ryun Chung \\
  \texttt{ho.chung@staff.uni-marburg.de} \\
  \AND
  {}\\[-2ex]
  Institute for Medical Bioinformatics and Biostatistics\\
  Philipps-Universität Marburg\\
  Hans-Meerwein-Straße 6, 35032 Germany\\
}

\begin{document}

\maketitle

\begin{abstract}

Computer-based decision systems are widely used to automate decisions in many aspects of everyday life, which include sensitive areas like hiring, loaning and even criminal sentencing. A decision pipeline heavily relies on large volumes of historical real-world data for training its models. However, historical training data often contains gender, racial or other biases which are propagated to the trained models influencing computer-based decisions. In this work, we propose a robust methodology that guarantees the removal of unwanted biases while maximally preserving classification utility. Our approach can always achieve this in a model-independent way by deriving from real-world data the asymptotic dataset that uniquely encodes demographic parity and realism. As a proof-of-principle, we deduce from public census records such an asymptotic dataset from which synthetic samples can be generated to train well-established classifiers. Benchmarking the generalization capability of these classifiers trained on our synthetic data, we confirm the absence of any explicit or implicit bias in the computer-aided decision. 

\end{abstract}

\section{Introduction}

Artificial intelligence (\textsc{ai}) finds extensive application in various classification tasks, ranging from buyer's guides to prioritizing \textsc{icu}-admissions and from hiring processes to self-driving cars. Computer-aided decision systems have demonstrated remarkable success in automating workflows and deriving accurate conclusions. However, it is important to recognize that the very factor contributing to the success of \textsc{ai} models also represents a potential vulnerability.

Any sufficiently complex machine-learning algorithm is expected to uncover all systematic patterns  inherent in the data to ensure realistic decision-making. 
This faithful representation of our social reality is essential, as it determines the practical utility of implementing \textsc{ai} processes in automating decision-making. 
On the other hand, faithfully generalizing from patterns and trends observed in real-world datasets automatically implies  replicating any discriminatory biases present within the dataset itself. 

In principle,  
two forms of discriminatory biases can be encountered in a classification setting. 
The first form is more apparent, enabling the identification of direct discriminatory relationships between a protected attribute, such as gender, and the final decision.
On the other hand, the second form is subtler, as it indirectly connects sensitive profiles to the decision through a discriminatory confounding with another predictor.
While the first form can be addressed by completely removing protected attributes from the dataset, the second form of bias is more challenging to detect and address. 
Most alarmingly, this second form of bias can resurface when the classifier generalizes to new data that persistently exhibits biases from society, even
 if offending confounding relationships have been correctly identified and removed during training.

As pattern-recognition and classification workflows in \textsc{ai}  become increasingly complex, it becomes more challenging to systematically identify and prevent both direct and indirect forms of discrimination influencing the computed-aided decision. This inability to guarantee the absence of known or suspected discriminatory biases hinders the broader application of \textsc{ai}, particularly in critical domains such as criminal sentencing or governance. In recent years, there has been a growing demand~\citep{Xu2021,mehrabi2021survey} for automation that is free from discriminatory biases, leading to the emergence of fair machine learning. Fair machine learning aims to accurately reproduce most patterns revealed by data while simultaneously restoring parity among sensitive profiles.

\begin{figure}
    \centering
\begin{tikzpicture}[scale=.92, remember picture ] 
\def\ymiddle{0}
\def\yfair{\ymiddle-2.5}
\node[] (data) at (-1,-0.2) {original data} ;
\node[draw,rounded corners=0.1cm] (empirical) at (2.1,-0.2) {empirical distro} ;
\node[draw] (pur) at (6., 0.2) {\textsc{pur} conditions};
\draw[-latex] (data)  -- (empirical) ;
\draw[-latex] (empirical) -- (pur);
\node[draw,rounded corners=0.1cm] (refdistro) at (6.,-1.1) {reference distro} ;
\draw[-latex,dotted] (empirical) |- (refdistro) ;
\node[draw,rounded corners=0.1cm] (Iproj) at (9.1, -0.5) {\textsc{pur} projection};
\draw[-latex] (pur) -| (Iproj);
\draw[-latex] (refdistro) -| (Iproj);
\node[] (classify) at (12., 0.5) {classify};
\draw[-latex,dashed] (Iproj) -- (classify); 
\node[] (synthetic) at (12., -1.5) {synthetic data};
\draw[-latex,dashed] (Iproj) -- (synthetic); 
\draw[-latex] (synthetic) -- (classify); 
\end{tikzpicture}
\caption{The flow chart of \textsc{pur} methodology which removes discriminatory biases to produce fair utility-driven  datasets. The latter can be used in training classifiers.}\label{fg:FlowChart}
\end{figure}
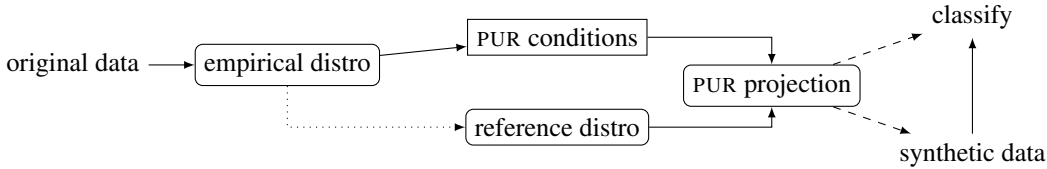

Within the context of fair machine learning, we adopt a systematic, model-independent approach that separates the task of de-biasing data from the actual training process of a classifier architecture.
This clear distinction allows us to provide robust mathematical assurances  of fairness on train and test data that are independent of the complexity of the model architecture. Figure~\ref{fg:FlowChart} illustrates our distinct approach to achieving fairness by appropriately modifying the data.

Given a real-world data
and after declaring protected predictors like gender, race/ethnicity, sexual orientation etc, one imposes a series of marginal constraints from the original data that any de-biased dataset has to obey, at least up to sampling noise.
We propose to require that our data fulfils demographic Parity, classification Utility and social Realism, in short \textsc{pur}. 
Starting from satisfying these rather intuitive constraints, we additionally demand to be as close as possible to the original data. In statistics, this optimization problem uniquely produces a fair probability distribution over profiles that precisely captures desired classifying relationships, while modifying (softly constrained) higher-order relationships to achieve demographic parity. 


In addition to drawing upon mathematical theorems, we demonstrate the logic and effectiveness of \textsc{pur} approach by concrete applications. 
Once we have derived a fair distribution from train data that summarizes real-world census records, we employ it as a natural classifier to make predictions on test data. This approach allows us to verify the absence of systematic bias against the designated protected attributes, while also confirming the classification utility of the natural classifier. Additionally, we leverage the fair distribution to generate synthetic datasets, which are then used to train random forests. 
This step highlights the ability of our methodology to generalize in broader contexts by augmenting established models. 

\section{Methodology}
\label{sc:Methodoloy}

In any classification setting, there minimally exists a --\,usually categorical\,-- feature, the so-called response variable $Y$ with at least two outcomes intimately related to a collection of explanatory features.
The latter are perceived as random variables comprising the set of predictors. Each predictor assumes an a priori different number of categories from some domain.\footnote{For compactness of notation, we use the same capital letter to collectively refer to the feature as well as its domain.}
Among predictors, we distinguish between \textit{protected} attributes $\mathbf S=\left(S_1,S_2,\ldots\right)$  that could entail \textit{sensitive} relationships to the response variable $Y$  and the remaining, \textit{unprotected} attributes $\mathbf X=\left(X_1,X_2,\ldots\right)$.
A tuple $(s_i, x_j)$ with $s_i\in S_i$ and $x_j\in X_j$ unambiguously characterizes then a predictor profile.

\subsection{Preliminaries}

\newcommand{\config}{(y, \mathbf s, \mathbf x)}


Any model-independent formulation of statistical problems necessarily relies on the \textit{joint} probability distribution $p$ over possible profiles $\config$ from the Cartesian product of response domain $Y$ with all predictor domains $\mathbf S$ and $\mathbf X$.
Armed with some estimate of this joint probability distribution, we can compute \textit{marginals} of selected features, say $Y$ and $X_i$  taking specific values $(y, x_i)$ by summing over all probabilities of joint profiles where the 
selected features
assume the specified values.
Determining marginal sums for all possible profiles in the Cartesian product of the selected domains defines in turn a marginal distribution.


A direct  estimate of joint probability distribution can be always obtained by calculating from the provided dataset relative frequencies $f\config$  which comprise the \textit{empirical} distribution.
Due to finite sample sizes or deterministic relationships like natural laws, not all profiles in the Cartesian product of feature domains are necessarily observed in real life, meaning that $f$ usually exhibits many sampling and structural zeros~\citep{bishop2007discrete}, respectively.   
In any case, we shall assume that all classes in $Y$ have been encountered in the data, at least once, as well as all sensitive profiles from $\mathbf S$. 

The theoretical machinery itself that is invoked in the next section is insensitive to the presence of zero estimates in the empirical distribution. 
However, to achieve fairness we shall make sure that 
any marginal $f(y,\mathbf s)$ involving the response to the sensitive attributes receives a finite probability. 
%
One straight-forward way to achieve this in probability space is via the pseudo-count method~\citep{MorcosE1293}:
\begin{equation}
\label{eq:regularized_empirical}
    f\config 
    \quad \rightarrow \quad
    \frac{f\config + \lambda/N}{1 + \vert Y\vert \vert \mathbf S\vert  \vert \mathbf  X\vert \lambda/N}
\end{equation}
$\vert\cdot\vert$ denotes the cardinality of feature domain. 
The hyper-parameter $\lambda$ which controls the regularization strength  was originally thought to be fixed to one. 
Nevertheless, our method robustly works with any $\lambda>0$. 
%

Since heavily extrapolating to unseen profiles could well be misleading, one could uniformly regularize the Cartesian product of all admissible labels $y\in Y$ with only the predictor profiles $(\mathbf s, \mathbf x)\in \mathbf S\times\mathbf X$ that have been observed in the data.  
Besides concerns~\citep{jaynes1968prior} regarding artifacts created by excessive regularization, assigning pseudo-counts to all joint profiles in $Y\times\mathbf S\times\mathbf X$ would quickly reveal the \textsc{np}-completeness underlying categorical problems with $L$ features  which scale at least as $2^L$.
By considering only predictor profiles that have been observed in real-world data (far below any bound posed by current computational technology), we are able to deduce exact results in Section~\ref{sc:FairDistro}.

\subsection{Problem statement}

The provided data could be --\,often severely\,-- biased against sensitive profiles  $\mathbf s\in\mathbf S$ corresponding to discriminated groups. 
Quantitatively, widely used~\citep{doi:10.1146/annurev.publhealth.27.021405.102103,mehrabi2021survey} measures of such \textit{disparity} are defined  as either ratios or differences between conditional probabilities. 
Focusing on a possible outcome $y\in Y$, we examine after marginalizing over $\mathbf X$,  the deviation of conditional $p(y\vert \mathbf s)$ given a sensitive profile $\mathbf s$ from a reference profile $\mathbf s_0$. The latter usually corresponds to a group which enjoys social privileges, also in accordance with the provided data.
Evidently, \textit{demographic parity} is restored whenever the conditional probabilities of the outcome become independent from protected attributes.

Generically, fair machine learning tries 
to avoid reproducing biased decisions against sensitive profiles that are advocated by training data. This objective appears to undermine the desired accuracy and generalization capability of a classification routine. 
In an extreme scenario, it would be possible to trivially create a fair classifier by assigning equal probability to every joint profile $\config$ 
at the expense of loosing any predictive power from the original data.
Already in previous work~\citep{9874243} on fair machine and representation learning, the notion has appeared of an ``optimal'' classifier  which partially compromises classification power to --\,almost\,-- achieve parity.

To be able to rigorously establish a definition of optimality, we first need to decouple the question about the architecture of a fair classifier from de-biasing training data.  
Focusing on the latter point, our scheme entirely operates at the level of (pre)-processing real-world datasets that are plagued by discriminatory biases. 
Ultimately, we want to  guarantee that the pre-processed data described by a joint distribution $p$ 
systematically satisfies parity among all profiles $\mathbf s\in\mathbf S$, while fully preserving real-world classification utility. 
As a result, any classifier would be at most exposed to training data described by $p$, instead of the original $f$ according to flow chart~\ref{fg:FlowChart}.

Translated in the language of distributions over joint profiles, our motivating goal becomes thus to find a \textit{fair} estimate for $p$ that enforces demographic Parity while retaining classification Utility of the original $f$. 
This amounts to requiring for all admissible profiles following marginal constraints:
\begin{itemize}
    \item demographic Parity
    \begin{equation}
    \label{eq:Parity}
        \sum_{\mathbf x\in\mathbf X} p(y, \mathbf s, \mathbf x)  = f(y) f(\mathbf s)
    \end{equation}
    \item decision Utility
    \begin{equation}
    \label{eq:Utility}
        \sum_{\mathbf s\in\mathbf s} p(y, \mathbf s, \mathbf x)  = f(y, \mathbf x)
    \end{equation}
    \item demographic Realism
    \begin{equation}
    \label{eq:Realism}
        \sum_{y\in Y} p(y, \mathbf s, \mathbf x)  = f(\mathbf s, \mathbf x)
    \end{equation}
\end{itemize}
Any $p$ that belongs to the convex set of distributions over $Y\times\mathbf S\times\mathbf X$ which satisfy these three groups of linear constraints in $p\config$ shall be called a \textsc{pur} distribution. 

In the \textsc{pur} scheme, demographic Parity is enforced as the absence of the correlation between response variable and sensitive attributes. Note that constraint~\ref{eq:Parity} implies $p(y \vert \mathbf s) = p(y) = f(y)$ for the derived conditional probabilities.
Consequently, any disparity ratio directly deduced from such \textsc{pur} distribution $p$ would be automatically one and any disparity difference zero. 

At the same time, decision Utility ensures that 
relationships  of the unprotected attributes $\mathbf X$ to the response variable $Y$ remain unaltered in $p$ and are not accidentally biased over pre-processing when correcting for Parity. 
Finally, demographic Realism prevents any form of indirect biasing (that could undermine our aim) by learning discriminatory relationships among predictors $\mathbf S$ and $\mathbf X$ which are currently present in society, as evidenced in the data.  

Below, we show via concrete applications that these  \textsc{pur} marginal conditions comprise a  minimal set of hard constraints required to systematically achieve our stated goals.
One of them is to stay as close as possible 
to the original dataset while correcting for any disparities. In terms of distributions, this can be expressed as minimization of the \textsc{kl} divergence~\citep{1056144,kullback1997information} from $f$,
\begin{equation}
\label{eq:KL_divergence}
    D_\textsc{kl}(p \vert \vert f) = \sum_{\config} p\config \log \frac{p\config}{f\config}
\end{equation}
over all joint distributions that fulfill constraints~\ref{eq:Parity},~\ref{eq:Utility} and~\ref{eq:Realism}. 

For the empirical $f$ as our \textit{reference} distribution, we have to use the regularized estimate~\ref{eq:regularized_empirical}, otherwise the \textsc{kl} divergence might not be well-defined especially for smaller datasets. 
Furthermore, we do not need to worry about unobserved profiles, as these have no information-theoretic impact, due to $0\cdot \log0 = 0$. Hence, the summation in Eq.~\ref{eq:KL_divergence} needs to go over the Cartesian product of the anticipated outcomes with observed-only predictor profiles.

\subsection{The fair solution}
\label{sc:FairDistro}
As it turns out~\citep{csiszar1975divergence,csiszar1991least}, 
under a consistent\footnote{Any constraint involving the prevalence of some joint profile $\config$ could render the linear system of coupled equations~\ref{eq:Parity}-\ref{eq:Realism} over-determined.} 
set of linear constraints,
the minimization of \textsc{kl} divergence in the probabilities over observed profiles 
poses a convex optimization problem. This always admits  a unique solution, the so-called \textit{information projection}~\citep{nielsen2018information} of empirical $f$ onto the convex solution space defined by constraints  \ref{eq:Parity},~\ref{eq:Utility} and~\ref{eq:Realism}, in short the \textsc{pur} projection of $f$.
In Appendix, we recapitulate  the proof of existence and uniqueness of the information-projection in a more applied fashion.

Generically, the information-projection of $f$ on the solution set defined by \textsc{pur} conditions would be a joint distribution 
with  real and not rational probabilities, the latter being relevant for finite sample size $N$.
Hence, one should think of the \textsc{pur} projection of $f$, signified by $q$, as the \textit{asymptotic} limit $N\rightarrow\infty$ at which a dataset with the Utility and Realism of the original data restores demographic Parity.
This is well demonstrated via sampling of counts from $q$. 

\paragraph{Production of synthetic data}
At finite sample size $N$, we can formally sample counts $Np\config\in\mathbb N_0$ from $q$ via the multinomial distribution  ${mult} (N p; q)$. 
In larger populations, it is permissible~\citep{hajek1960limiting} to use the multinomial instead of the formally more appropriate hyper-geometric distribution to sample datasets that are smaller than the population size. 
Incidentally, this sampling operation  provides a coherent way to generate synthetic data described by $p$ that differ from \textsc{pur} projection  by mere sampling noise.

In other words, synthetic data produced from $q$ as indicated by the last step in Figure~\ref{fg:FlowChart} would not introduce any systematic demographic bias against $\mathbf S$, as long as this had been fully removed from $q$.
Indeed, a large-$N$ expansion,
\begin{equation}
   \log {mult} (N p; q) = - N D(p \vert\vert q) + \dots
\end{equation}
best demonstrates (recall that $D(p \vert\vert q)\rightarrow 0$ iff $ p \rightarrow q$) how synthetic datasets sampled from \textsc{pur} projection $q$ become with increasing $N$ more and more concentrated around it. 

\paragraph{Alternative reference distribution}
As argued below Eq.~\ref{eq:KL_divergence}, an intuitive reference distribution to select the \textsc{pur} projection  is the regularized empirical distribution.
By tuning $\lambda$ in Eq.~\ref{eq:regularized_empirical}, we can always bring the regularized $f$ closer to the uniform distribution $u$ which assigns the same probability to any joint profile $\config$. In the limit of $\lambda \rightarrow \infty$, we uncover due to ($H$ denotes Shannon's entropy)
\begin{displaymath}
    H[p] = - D_\textsc{kl}(p \vert \vert u) + \log\left( \vert Y\vert \vert \mathbf S\vert  \vert \mathbf  X\vert\right)
\end{displaymath}
the principle of Maximum entropy~\citep{jaynes2003probability}, in short \textsc{m}ax\textsc{e}nt under \textsc{pur} constraints. 

To avoid disclosing higher-order effects between predictors and response, an aspect of paramount importance in privacy-related applications, one could well consider the \textsc{pur} projection of $u$
as starting point for fair model-building. Such choice goes in the direction of~\citep{celis2020data}, though in our setup we ensure that the optimal \textsc{m}ax\textsc{e}nt distribution exactly satisfies the fairness constraints~\ref{eq:Parity}-\ref{eq:Realism}.  
As the proposed formalism remains structurally the same under any reasonable (i.e\ not unjustifiably biased) reference distribution in Eq.~\ref{eq:KL_divergence},  it bears the potential to be readily applied in the cross-roads of fair and private machine learning, in the spirit of~\citep{chaudhari2022fairgen,pujol2022prefair}. 

\paragraph{The iterative minimization of information divergence}
After receiving a dataset described by empirical distribution $f$ and having decided about a reference distribution, most intuitively $f$ itself, 
we need to compute its unique \textsc{pur} projection.
In most cases, there exists no closed-form solution, so that some iterative method must be invoked. 
In principle, multi-dimensional Newton-based methods could quickly find $q$ starting e.g.\ from $f$,  after reducing \textsc{pur} conditions to linearly independent constraints~\citep{loukas2022model}. 

Another class of iterative approaches which is particularly tailored to enforce marginal constraints on reference distribution 
is the Iterative Proportional Fitting (\textsc{ipf}) algorithm, first introduced in~\citep{kruithof1937telefoonverkeersrekening}.  
As it is argued in~\citep{ireland1968contingency,darroch1972generalized,haberman1974log} and rigorously shown in~\citep{csiszar1975divergence,bishop2007discrete}, this iterative scheme has all the guarantees (see also discussion~\citep{phdthesis}) to converge to the \textsc{pur} distribution within the desired numerical tolerance.
At the practical level, one can directly work with the redundant set of conditions~\ref{eq:Parity}-\ref{eq:Realism} (e.g.\ both marginals $p(y,\mathbf x)$ and $p(y,\mathbf s)$ imply the prevalence $p(y)$) manifestly preserving  interpretability.  

Programmatically, 
we start from $p^{(0)} = f$ and iteratively update our running estimate by imposing \textsc{pur} conditions, 
\begin{align*}
    p^{(n+1)}\config =&\,\, p^{(n)}\config \frac{f(y) f(\mathbf s)}{p^{(n)}(y,\mathbf s)}\\
    p^{(n+2)}\config =&\,\, p^{(n+1)}\config \frac{f(y, \mathbf x)}{p^{(n+1)}(y,\mathbf x)}\\
    p^{(n+3)}\config =&\,\, p^{(n+2)}\config \frac{f(\mathbf s, \mathbf x)}{p^{(n+2)}(\mathbf s, \mathbf x)}
\end{align*}
until $p^{(n)}\rightarrow q$ within numerical tolerance. Note that the order with which we impose marginal constraints does not influence the eventual convergence, as long as it remains fixed throughout the procedure. 
Besides general-purpose \textsc{ipf} packages~\citep{IPFpypi} and~\citep{IPFr} for Python and \texttt{R}, we provide in supplementary material a self-contained data-oriented implementation of \textsc{ipf} routine based on \texttt{numpy} and \texttt{pandas} modules~\citep{2020NumPy-Array,mckinney2010data}.

\section{Application}

To demonstrate the efficiency and flexibility of the developed methodology we consider census data from the USA.

\subsection{Multi-label classification}

In the period from 1981 to 2013, there exist census records publicly available under~\url{ https://www.kaggle.com/fedesoriano/gender-pay-gap-dataset}.
After appropriate binning in lower, mid-range and higher salaries, 
we choose the response variable $Y = \texttt{hourly salary ranges}$ with five outcomes.
From the provided raw data, it is straight-forward to define a predictor profile by unprotected attributes $\mathbf X = \left(\texttt{age group, education degree, occupation sector} \right)$ and protected attributes $\mathbf S = \left(\texttt{gender, race} \right)$.
As sensitive profiles, we examine
$\texttt{gender}=\texttt{male,\,female}$ and $\texttt{race}=\texttt{white,\,black,\,hispanic}$.

Furthermore, we use the empirical $f$ associated to each census year appearing in the raw data to sample~\citep{politis1999subsampling} a wealth of training and test datasets within the original sample sizes $\sim35'000-55'000$ entries. 
Details on the statistics of relevant features and the defined census profiles as well as on generation of train-test data are given in Appendix.

As a measure of unfairness, we choose to look at \textit{attributable disparity} defined as (cf.~\citep{walter1976estimation})
\begin{equation}
\label{eq:AttributableDisparity}
    p(y\vert \mathbf s) - p(y\vert \mathbf s_0)
\end{equation}
w.r.t.\ some reference group who enjoyed social privileges at the time of the survey.
A quick inspection of empirical statistics for $p=f$ reveals that $\mathbf s_0 =\left(\texttt{male,\,white}\right)$ had a conditional probability of roughly below 50\% to earn up to 20\$, as opposed to all other sensitive profiles 
with the conditional probability in the lower salary range rising above $90\%$ for $\mathbf s=\left(\texttt{female}, \texttt{hispanic}\right)$.
 The picture gets reversed for higher salaries.
 Evidently, \ref{eq:AttributableDisparity} vanishes identically whenever $p=q$, where by construction fair  $q$ denotes the $\textsc{p}(\text{ur})$-projection of a train distribution.

\begin{figure}
    \centering
    \includegraphics[width=1\linewidth]{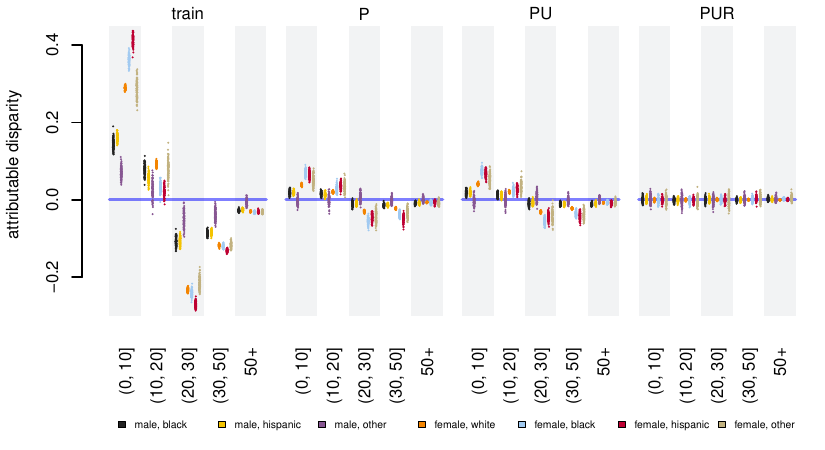}
    \caption{Attributable disparity over salary classes estimated by the prediction on simulated test data of information projection  of train distributions under various conditions. 
    Blue line denotes the estimate $p_\textrm{pred}(y\vert \mathbf s_0)$ used as reference in~\ref{eq:AttributableDisparity}. The original data refers back to 1981.}
    \label{fig:fair-1981}
\end{figure}

Similar to the original observations made in~\citep{10.1257/jel.20160995}, the positive and negative disparity slowly approaches zero over the years in both lower and higher salaries, respectively. 
Still, there  remains up to 2013  a significant amount of demographic disparity~\ref{eq:AttributableDisparity} up to 30\%  over the whole salary range. %
Sampled from the original empirical distributions of the different years, both train and test data exhibit similar trends reproducing in particular the discriminatory bias.
Indeed, this can be easily confirmed by plotting the average attributable disparity  alongside its fluctuation scale over simulated  train data, see first column of Figure~\ref{fig:fair-1981}.

\paragraph{Generalization and Parity}
Henceforth we focus on year 1981; an analogous analysis and exposition of results for following census years is provided in Supplementary Material.
After running \textsc{ipf} to incorporate all \textsc{pur} conditions stemming from (mildly regularized with $\lambda=10^{-4}$) empirical distributions 
describing the simulated train data, we obtain their \textsc{pur}  
projections $q$.
In principle, we could use the \textsc{pur}  
projections to produce a wealth of synthetic data points and subsequently train more elaborate classifiers 
to perform predictions on test data. 
Nevertheless, there is nothing that prevents us from using $q$ itself 
as a natural classifier according to the fundamental rule of conditional probabilities:
\begin{equation}
\label{eq:PredictedDistro}
    p_\textrm{pred}\config = q(y \vert \mathbf s, \mathbf x) \cdot f_\textrm{test}(\mathbf s, \mathbf x)
\end{equation}
where $f_\textrm{test}$ denotes the empirical distribution of simulated test data.

Beyond mere intuition, to illustrate the necessity of all \textsc{pur} conditions  we determine using \textsc{ipf} the information projection of each train data under Parity (\textsc{p}), Parity and Utility (\textsc{pu}) and eventually \textsc{pur}. The average predictions (alongside the scale of fluctuations over simulated data) of different combinations of conditions are depicted in the three last columns of Figure~\ref{fig:fair-1981}, respectively. 

Clearly, demographic Parity is systematically (i.e.\ beyond mere sampling noise) achieved only using the \textsc{pur} projection as a natural classifier on test data.
In particular, demographic Realism enables $q$ to compensate for test data discriminating against sensitive groups through e.g.\ lower prevalence in highly paid jobs. 
It is however noteworthy that 
minimizing the \textsc{kl}-divergence from the train empirical distribution under Parity condition alone still improves the situation compared to directly using 
the train distribution itself as a classifier, cf.\ first two columns of Figure~\ref{fig:fair-1981}.

\begin{figure}
    \centering
    \includegraphics[width=0.6\linewidth]{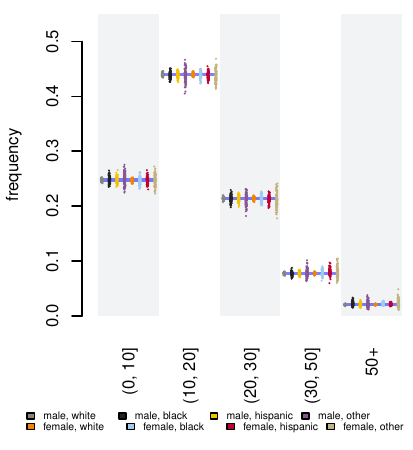}
    \caption{Natural predictor $p_\textrm{pred}(y\vert \mathbf s)$ of the \textsc{pur}-projection of train distribution 
    evaluated on simulated test data for all profiles in $\mathbf S$.
    Blue line denotes $f(y)$ as computed from the empirical distribution of the original data.}
    \label{fig:PUR-abs}
\end{figure}

To closer illustrate the situation described by the \textsc{pur} projection, we compare in Figure~\ref{fig:PUR-abs} the conditional probability  $p_\textrm{pred}(y\vert \mathbf s)$ for all seven $\texttt{gender} \times \texttt{race}$ profiles against the original marginal $f(y)$.  
Evidently, \textsc{pur} predictions obey the general variability in the empirical distribution of salaries $f(y)$ --\,triggered by e.g.\ different occupations and education levels in $\mathbf X$. 
As anticipated, the conditional estimate of~\ref{eq:PredictedDistro} over simulated samples statistically fluctuates around this global profile without any systematic discriminatory tendency triggered by $\mathbf S$.

\paragraph{Generalization and Utility}
For any machine-learning algorithm, a measure of its generalization capability is required. In the given context of fairness, where data has been deliberately --\,albeit in a controlled manner\,-- 
modified, the \textsc{kl} divergence of predicted joint distribution~\ref{eq:PredictedDistro} from the empirical distribution of test data
could become mis-leading.
On the contrary, it is most natural to introduce a Utility-based  metric to quantify generalization error in a model-independent way. Our suggestion is the \textsc{kl} divergence of the test $Y-\mathbf X$ marginal from the corresponding predicted marginal: 
\begin{equation}
\label{eq:Generalization_Utility}
    \sum_{y\in Y}\sum_{\mathbf x\in\mathbf X} f_\texttt{test}(y, \mathbf x) \log \frac{f_\textrm{test}(y, \mathbf x)}{p_\textrm{pred} (y, \mathbf x) }
\end{equation}

Self-consistently, the metric becomes zero by merit of condition~\ref{eq:Utility}, when replacing the test with the train empirical distribution and the  predicted distribution~\ref{eq:PredictedDistro} with the \textsc{pur} projection of train distribution. 

\begin{figure}
    \centering
    \includegraphics[width=0.55\linewidth]{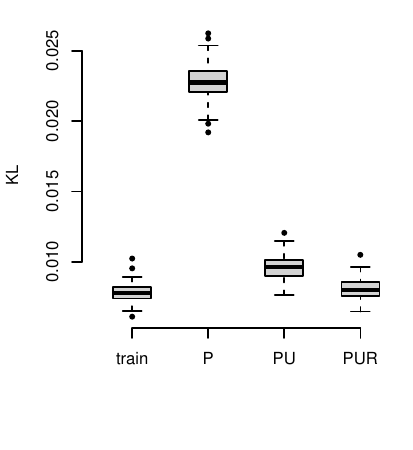}
    \caption{Utility-based generalization error~\ref{eq:Generalization_Utility} on simulated test data for the natural classifier associated to the information-projection under  different combinations of conditions~\ref{eq:Parity}-\ref{eq:Realism}.}
    \label{fig:utility-1981}
\end{figure}

In Figure~\ref{fig:utility-1981}, we give a box plot for the Utility-based generalization metric. 
Within the scale of variation of the simulated datasets, we can safely conclude that the natural classifier constructed out of the \textsc{pur} projection of train data performs in average as good as using the train distribution itself, cf.\ first and last columns. 
Similar dispersion diagrams over salary classes and box plots for all methods  are listed in Supplementary Material regarding all census years.

\subsection{Binary classification}

A classification task performed on the adult dataset \url{https://archive.ics.uci.edu/ml/datasets/adult} provides additional support for the importance of implementing all \textsc{pur} conditions~\ref{eq:Parity}-\ref{eq:Realism} in order to achieve demographic Parity.
Here, the response variable $Y$ is the yearly income, which is either
high ($> 50k$) or low ($\leq 50k$). As before, protected attributes $\mathbf S$  are \texttt{gender} and \texttt{race}; the latter also binarized in \texttt{white} or \texttt{non-white}.
Finally, the unprotected attributes $\mathbf X$ are $\texttt{age group} \in \lbrace
\texttt{young, middle, senior}\rbrace$, 
$\texttt{workclass} \in \lbrace\texttt{gov, private, self-employed}\rbrace$ and $\texttt{education} \in \lbrace\texttt{dropout, highschool, above-highschool}\rbrace$.

After splitting the original dataset in train and test data, we compute the relevant marginals~\ref{eq:Parity}-\ref{eq:Realism} from $f_\textrm{train}$ in order to derive the information projection of train distribution $f_\textrm{train}$ under Parity and under all \textsc{pur} conditions, the $\textsc{p}$- and $\textsc{pur}$-projection of $f_\textrm{train}$ respectively.
Following our flowchart~\ref{fg:FlowChart}, we subsequently generate  a wealth of synthetic datasets from the \textsc{p(ur)}-projections in order to train random forest classifiers  on them using module~\citep{scikit-learn}. 
In addition, we provide analogous results for the preivacy-relevant \textsc{m}ax\textsc{e}nt distribution under \textsc{pur} conditions. All details and code for data generation and training are given in Supplementary Material.

\begin{figure}
    \centering
    \includegraphics[clip, trim=1cm 9cm 1cm 8cm, width=1\linewidth]{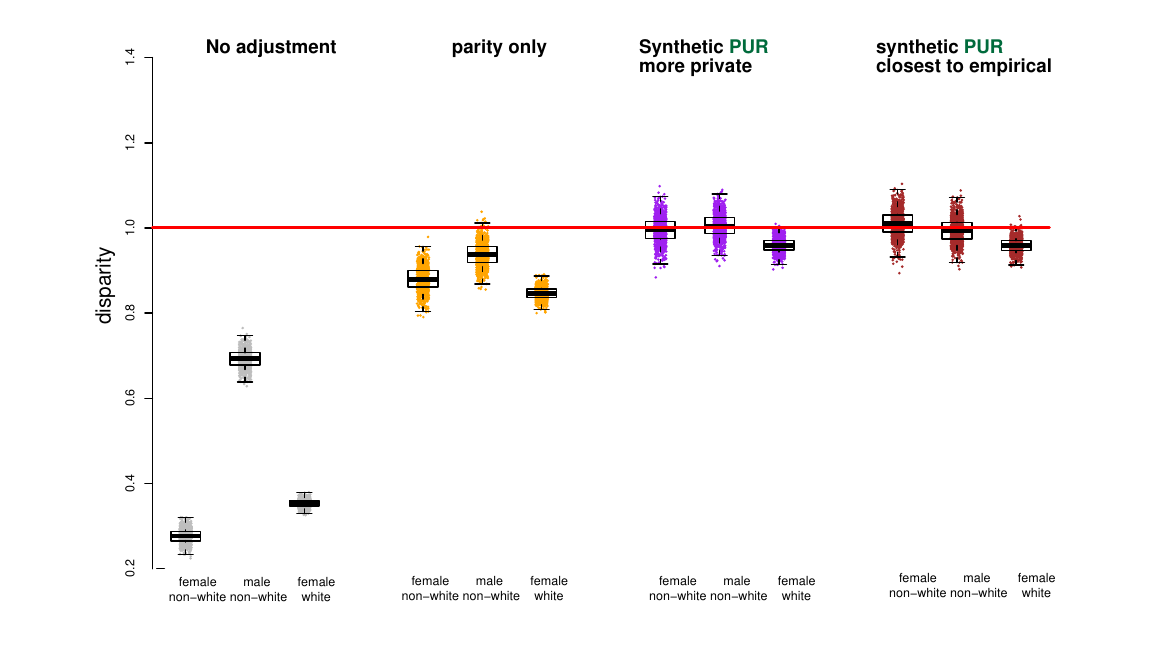}
    \caption{Disparity ratio estimated by the prediction on the same test data  
    of Random Forest Classifiers trained on synthetic datasets generated from the \textsc{p(ur)}-projection of $f_\textrm{train}$ and $u$.}
    \label{fig:bin-classification}
\end{figure}

As evidenced from the first column in Figure~\ref{fig:bin-classification}, 
random forests trained on 
synthetic data 
generated from $f_\textrm{train}$ 
without adjusting for Parity reproduce via their predictions the
biases in the adult dataset. This means that  sensitive profiles $\mathbf s =(\texttt{female}, \texttt{non-white})$, $(\texttt{male}, \texttt{non-white})$  and $(\texttt{female}, \texttt{white})$ 
with high income occur much less often than $\mathbf s_0=(\texttt{male}, \texttt{white})$ with high income.
In binary classification, this observation easily translates
into a ratio of conditionals as measure of disparity, i.e.\ the fraction of individuals with high income in the discriminated groups versus the
privileged group $\mathbf s_0$; in all three discriminated groups this fraction is below 80\% without further adjustments. 

A Random Forest Classifier trained on synthetic data generated by the $\textsc{p}$-projection of $f_\textrm{train}$ reintroduces discriminatory bias
when predicting on test cases -- albeit not so strong as in the unadjusted case. This bias is mediated 
via discriminatory correlations in test data between unprotected and protected  attributes, since the $\textsc{p}$-projection 
does not adhere to demographic Realism.
On the contrary,  Random Forest Classifiers trained on synthetic data generated
by both \textsc{pur}-distributions of $f$ and of $u$ remain de-biased up to generalization error, when evaluated on test data, cf.\ last two columns of Figure~\ref{fig:bin-classification}. 
This confirms the necessity and adequacy of the \textsc{pur} scheme.




%
%

\pagebreak

\appendix
\begin{center}
    \textbf{Appendix \& Supplementary Material}
\end{center}

\newcommand{\idx}{\alpha}

\DeclarePairedDelimiterX\infdivx[2]{(}{)}{
  #1\;\delimsize\|\;#2
}
\newcommand{\infdiv}{D_\textsc{kl}\infdivx}

\section{Theory}
\label{sc:theory}

In the main paper, we have investigated relationships between some multi-label response variable $Y$ and protected $\mathbf S=S_1, S_2,\ldots$ as well as unprotected $\mathbf X = X_1, X_2,\ldots$ attributes.
When addressing fairness in a model-independent manner, any question unavoidably deals with probabilities over social profiles that live in the Cartesian product of 
\begin{equation}
    Y\times \mathbf S\times \mathbf X \equiv 
    Y\times S_1\times S_2 \times \dots \times X_1\times X_2\times \dots 
    ~.
\end{equation}
To effectively de-bias a given dataset, it suffices to formally handle attributes as categorical variables by imposing marginal constraints on the probability simplex. Hence, we refrain from discussing more general forms of linear constraints.

Primarily, we are  interested in producing a demographically fair version of the social phenomenology appearing in a given dataset that still retains phenomenological relevance for present society. 
Within the model-independent formulation, phenomenology is expressed as a system of linear equations.
Starting point of \textsc{pur} methodology are thus three sets of marginal constraints 
\begin{align}
\label{eq:PhenoConstraints}
    p(y, \mathbf s) =&\,\, \sum_{\mathbf x\in\mathbf X} p(y, \mathbf s, \mathbf x)  \overset{!}{=} f(y) f(\mathbf s)
    \quad,
    \\[1ex]
    p(y, \mathbf x) =&\,\,\sum_{\mathbf s\in\mathbf s} p(y, \mathbf s, \mathbf x)  \overset{!}{=} f(y, \mathbf x)
    \quad\text{and}\quad
    p(\mathbf s, \mathbf x) =\sum_{y\in Y} p(y, \mathbf s, \mathbf x)  \overset{!}{=} f(\mathbf s, \mathbf x)
    \nonumber
\end{align}
imposed on joint probability distributions $p$ over social profiles to achieve demographic Parity, while intuitively incorporating Utility and Realism, respectively.
The shorthand notation $\mathbf x\in\mathbf X$ means $x_1\in X_1$, $x_2\in X_2,\ldots$ 
We shall refer to the convex subspace of the probability simplex over  $Y\times\mathbf S\times\mathbf X $ which incorporates all those distributions that satisfy our aims by \textsc{pur}:
\begin{equation*}
    p\in\textsc{pur} \quad\Leftrightarrow\quad p \text{ satisfies }~\eqref{eq:PhenoConstraints}\,.
\end{equation*}

\subsection{The optimization program}
\label{ssc:Iprojection}

To illustrate the linear character of phenomenological problem at hand, 
we choose to arbitrarily enumerate profiles in the Cartesian product  
via $enum:Y\times\mathbf S\times\mathbf X\rightarrow \mathbb N$.
For compactness, we denote   $\idx \equiv enum\config\in\{1,\ldots,\vert Y\vert \vert \mathbf S \vert \vert \mathbf X\vert\}$ where shorthand notation  $\vert \mathbf S\vert = \vert S_1\vert \vert S_2\vert \cdots$ and  $\vert \mathbf X\vert = \vert X_1\vert \vert X_2\vert \cdots$ is understood for the cardinalities of protected and unprotected attributes, respectively.
%
Correspondingly, we enumerate marginal profiles  by the maps
\begin{align*}
    &enum_{P}(y, \mathbf s) \in \{1,\ldots, \vert Y\vert \vert \mathbf S \vert\}
    \quad,\quad  
    enum_{U}(y, \mathbf x) \in \{ \vert Y\vert \vert \mathbf S \vert  , \ldots,  \vert Y\vert \vert \mathbf S \vert + \vert Y\vert \vert \mathbf X \vert \}~,
    \\[1ex]
    &enum_{R}(\mathbf s, \mathbf x) \in \{ \vert Y\vert \vert \mathbf S \vert + \vert Y\vert \vert \mathbf X \vert,\ldots,  \vert Y\vert \vert \mathbf S \vert + \vert Y\vert \vert \mathbf X \vert + \vert S\vert \vert \mathbf X \vert \equiv D\}&
\end{align*}
which we collectively signify  by $m\in\{1,\ldots, D\}$.
A column vector with elements $f_m$ facilitates then  all  empirical moments appearing in Eq.~\eqref{eq:PhenoConstraints}, $f(y) f(\mathbf s)$, $f(y,\mathbf x)$ and $f(\mathbf s,\mathbf x)$ .

The linear-algebraic character of a marginal sum can be well demonstrated via a binary coefficient matrix $\mathbf C$ operating on probabilities to map them onto marginals. 
In terms of  $\mathbf C$, we can write the \textsc{pur} constraints
as a redundant, linear system of $D$ coupled equations 
\begin{equation}
    \label{eq:LinearSystem}
    \sum_{\idx=1}^{M} C_{m,\idx}\,p_\idx = f_m
\end{equation}
in generically $M\equiv \vert Y\vert \vert \mathbf S \vert \vert \mathbf X\vert$  variables --\,the probabilities $p_\idx\in[0,1]$.
In this language, we are concerned with non-negative vectors in $\mathbb R^M$ --\,representing distributions on the simplex--\, that solve linear system~\eqref{eq:LinearSystem}.
Any elementary row operation on $\mathbf C$ gives a phenomenological problem which is equivalent to~\eqref{eq:PhenoConstraints}. 

Any structural or sampling zero (due to deterministic or finite-$N$ behavior, respectively) must be considered separately~\cite{bishop2007discrete}.
The former type of zero probabilities is a consequence of logic and natural laws, hence such probabilities can be immediately set to zero. Obviously, any form of regularization should avoid re-introducing them later. 
The latter form of zero probabilities could be trickier to uncover. Besides regularization schemes suggested in the main paper, 
any empirical marginal that vanishes implies due to non-negativity that all probabilities entailed in the marginal sum must be also zero:
\begin{equation}
    f_m = 0 \quad\Rightarrow\quad
    C_{m,\idx}p_\idx = 0 ~~(\text{no sum})~.
\end{equation}
Such constraints reduce both the number of stochastically active profiles (columns of $\mathbf C$) as well as the number of non-trivial marginal constraints (rows of $\mathbf C$). 
We shall refer to the resulting coefficient matrix as the reduced form of $\mathbf C$.

The rank of the reduced coefficient matrix 
defines the linearly independent constraints implied by the linear problem 
independently of the particular parametrization of non-zero marginals. 
Evidently, linear system~\eqref{eq:LinearSystem} admits at least one non-negative solution, the empirical distribution $f$ itself. 
As long as the rank of the reduced coefficient matrix remains smaller than the number of its columns, 
there exist due to Rouché–Capelli theorem infinitely many,  non-negative by continuity  solutions.

\paragraph{The information projection}
One crucial fact is the existence and uniqueness of a distribution $q$ which 
satisfies all phenomenological constraints \eqref{eq:LinearSystem} while staying closest to a sensible reference distribution $q^{(0)}$. \
In the context of fair-aware machine learning, we have argued that such reference could either be a regularized version of the empirical distribution $f$ or the uniform distribution $u$ over admissible social profiles.
Conventionally, $q$ is called the information projection of $q^{(0)}$ on the \textsc{pur} subspace of the simplex, for us in short the \textsc{pur} projection. 
Mathematically, the \textsc{pur} projection satisfies 
\begin{equation}
\label{eq:InformationProjection}
    D_\textsc{kl}(q\vert\vert q^{(0)}) \leq D_\textsc{kl}(p \vert\vert q^{(0)}) \quad \forall \, p \in \textsc{pur}~.
\end{equation}
We emphasize that $q^{(0)}$ does not need to belong to 
\textsc{pur} space --\, and in fact it would not, otherwise our society would be exactly  fair, at least from the demographic perspective. 
%

The uniqueness of a minimum of \textsc{kl} divergence $D_\textsc{kl}(p \vert\vert q^{(0)})$ immediately follows in probability space by the convexity of the feasible region of the phenomenological problem~\eqref{eq:PhenoConstraints} at hand; combined with strict 
convexity~\cite{cover2012elements} of the  \textsc{kl}-divergence in its first argument thought as a function $[0,1]^M\rightarrow\mathbb R_0^+$.
Hence,  the \textsc{kl} divergence possesses one global minimum in \textsc{pur} subspace, at most.
Regarding joint distributions as column vectors in $[0,1]^M$ naturally represents the \textsc{pur} subspace by a non-empty, convex, bounded and closed --\,hence compact\,-- subset of $[0,1]^M$, viz.\ \eqref{eq:LinearSystem}, over which any continuous function necessarily attains a minimum by the extreme value theorem. In total, we conclude that the \textsc{kl} divergence must attain its global minimum in the \textsc{pur} subspace.

\subsection{Iterative proportional fitting}
\label{ssc:IPF}

In fair-aware applications, we have advocated the use of \textsc{ipf} algorithm to obtain the information projection that satisfies empirical marginal constraints starting from $p^{(0)}=q^{(0)}$. 
If $p_\idx=0$ as either a structural or a sampling zero, then the algorithm has trivially converged to it already at the first iteration. 
Using the linear-algebraic characterization, we can succinctly write in terms of the coefficient matrix the update rule for the stochastically interesting probabilities after $n$ fittings onto all positive marginals $f_m$, 
\begin{equation}
\label{eq:IPF_update}
 p^{(nD+m)}_\idx = 
 p^{(nD+m-1)}_\idx \left( \frac{f_m}{p^{(nD+m-1)}_m} \right)^{C_{m,\idx}}
\quad\forall\,\idx=1,...,M
~.
\end{equation}

Now, we show~\cite{csiszar1975divergence} that \textsc{ipf} in this setting converges to the \textsc{pur} projection. 
First, we need to verify that the iterative algorithm converges to a distribution within the \textsc{pur} subspace. 
For any probability distribution $p$  satisfying the given set of linear constraints~\eqref{eq:LinearSystem}, 
 the relation 
\begin{equation}
\label{eq:IPF:Divergences_Relation}
\infdiv{ p}{ p^{(nD+m-1)}} = \infdiv{ p}{ p^{(nD+m)}} + \infdiv{ p^{(nD+m)}}{ p^{(nD+m-1)}}
\end{equation}
holds.\footnote{Note that all \textsc{kl} divergences remain finite due to $0\cdot\log0 = 0$, as long as the reference distribution does not assume any zero probabilities for profiles that are later observed in the data. } 
This relation directly follows from 
\begin{equation*}
\sum_{\idx=1}^M \left[ p_\idx -  p^{(nD+m)}_\idx\right]\log\frac{ p^{(nD+m)\phantom{-1}}_\idx}{ p^{(nD+m-1)}_\idx}  = 
\log \frac{f_m}{p^{(nD+m-1)}_m} 
\sum_{\idx=1}^M C_{m,\idx}\left[ p_\idx -  p^{(nD+m)}_\idx\right]
= 0
~, 
\end{equation*}
after substituting update rule~\eqref{eq:IPF_update} 
whose form automatically ensures that
\begin{equation}
\label{eq:IPF:ath_marginal}
p^{(nD+m)}_m = \sum_{\idx=1}^M C_{m,\idx} p^{(nD+m)}_\idx = f_m
\end{equation}
after fitting onto the $m$-th marginal, so that each term vanishes identically in the latter sum given $p\in\textsc{pur}$.

After $n$ cycles, it follows from \eqref{eq:IPF:Divergences_Relation} by induction 
\begin{equation*}
\infdiv{ p}{ p^{(0)}} - \infdiv{ p}{ p^{(nD)}} = 
\sum_{n'=0}^{n-1}\sum_{m=1}^D\infdiv{ p^{(n'D+m)}}{ p^{(n'D+m-1)}}
~.
\end{equation*}
Since the difference on l.h.s.\ stays finite as $n \rightarrow\infty$, the series over non-negative terms on r.h.s.\ would be finite, as well. 
By the Cauchy criterion, there must exist for any $\varepsilon>0$ some $n^*\in\mathbb N$ so that 
\begin{equation}
\infdiv{ p^{(nD+m)}}{ p^{(nD+m-1)}} < \varepsilon 
\quad\text{for}\quad n\geq n^* \quad\text{and}\quad m=1,...,D
~.
\nonumber
\end{equation}
In turn, this implies that $p^{(nD+m)}$ induces a Cauchy sequence, thus establishing the existence of a generically real-valued limiting distribution  $q'$. 
Because each $ p^{(nD+m)}$   fulfills the $m$-th marginal sum, viz.\ Eq.~\eqref{eq:IPF:ath_marginal}, cycling through all marginals $m=1,...,D$ 
forces the limiting distribution $q'$ to satisfy them all. Consequently, the limiting distribution $q'$ has to belong to \textsc{pur}. 

In particular, we conclude  after  finitely many steps that
\begin{equation}
\hspace{1.3cm} p^{(nD+m-1)} \approx  p^{(nD+m)}
\quad\text{for}\quad n\geq n^* \quad\text{and}\quad  m=1,...,D
\nonumber
\end{equation}
within the desired tolerance $\varepsilon$ (dictated e.g.\ by machine precision), which is obviously of practical importance. 
In cases, when \textsc{ipf} fails to converge sufficiently fast within the desired tolerance, one can resort to its generalizations,  approximations based on gradient descend or Newton-based routines (see main text for references therein).

Eventually, it remains to verify that $q'$ is indeed the \textsc{pur} projection. 
Given two distributions $ p, \tilde p\in\textsc{pur}$, it can be inductively shown that
\begin{equation}
\label{eq:IPF:ith_Pythagorian}
\sum_{\idx=1}^M \left[ p_\idx - \tilde p_\idx\right] \log  \frac{p^{(nD+m)}_\idx}{q^{(0)}_\idx} = 0
\quad\text{for}\quad n=0,1,2,... \quad\text{and}\quad m=1,...,D
~. 
\end{equation}
Using \textsc{ipf} update rule \eqref{eq:IPF_update} we can indeed break the estimate at $nD+m+1$  into two parts:
\begin{align*}
\sum_{\idx=1}^M \left[ p_\idx -  \tilde p_\idx\right] \log \frac{p^{(nD+m+1)}_\idx}{q^{(0)}_\idx}
=&\,\,
\sum_{\idx=1}^M  \left[ p_\idx -  \tilde p_\idx\right] \log  \frac{p^{(nD+m)}_\idx}{q^{(0)}_\idx}
\\
&~~+
\log\frac{f_m}{p^{(nD+m)}_m}\sum_{\idx=1}^M  C_{m+1,\idx}\left[ p_\idx -  \tilde p_\idx\right]
= 0
~.
\end{align*}
The second summation vanishes identically, since  both $ p$ and $\tilde p$ reproduce the observed $m$-th marginal from $ f$ (otherwise they would not belong to \textsc{pur}).
At the same time, the first summation is zero by the inductive assumption. 
Starting from $n=0$ and $m=0$ the vanishing of the first summation is trivial for $p^{(0)} = q^{(0)}$,
thus verifying the induction.

Finally, taking $n \rightarrow \infty$ in  Eq.~\eqref{eq:IPF:ith_Pythagorian} and setting $ \tilde p=q'\in\textsc{pur}$ (as concluded above) results into
\begin{equation*}
\sum_{\idx=1}^M \left[ p_\idx - q'_\idx\right] \log \frac{q'_\idx}{q^{(0)}_\idx} = 0
\quad\Leftrightarrow\quad
\infdiv{p}{q^{(0)}} = \infdiv{p}{q'} + \infdiv{q'}{q^{(0)}}
~.
\end{equation*}
Since the \textsc{kl} divergence is non-negative definite, it directly follows 
$\infdiv{p}{q^{(0)}} \geq  \infdiv{q'}{q^{(0)}}$.
Equation~\ref{eq:IPF:ith_Pythagorian} was shown for 
arbitrary distributions $ p\in \textsc{pur}$.
Consequently, we conclude from definition~\eqref{eq:InformationProjection} of the information projection and its uniqueness that   
$q'$ is indeed the \textsc{pur} projection, namely $q'=q$. 
This formally shows that \textsc{ipf}  converges to the information projection of reference  distribution onto the \textsc{pur} subspace.

\section{Applications}

\subsection{The gender-ethnicity gap}

\begin{figure}[t]
    \centering
    \includegraphics[scale=0.3]{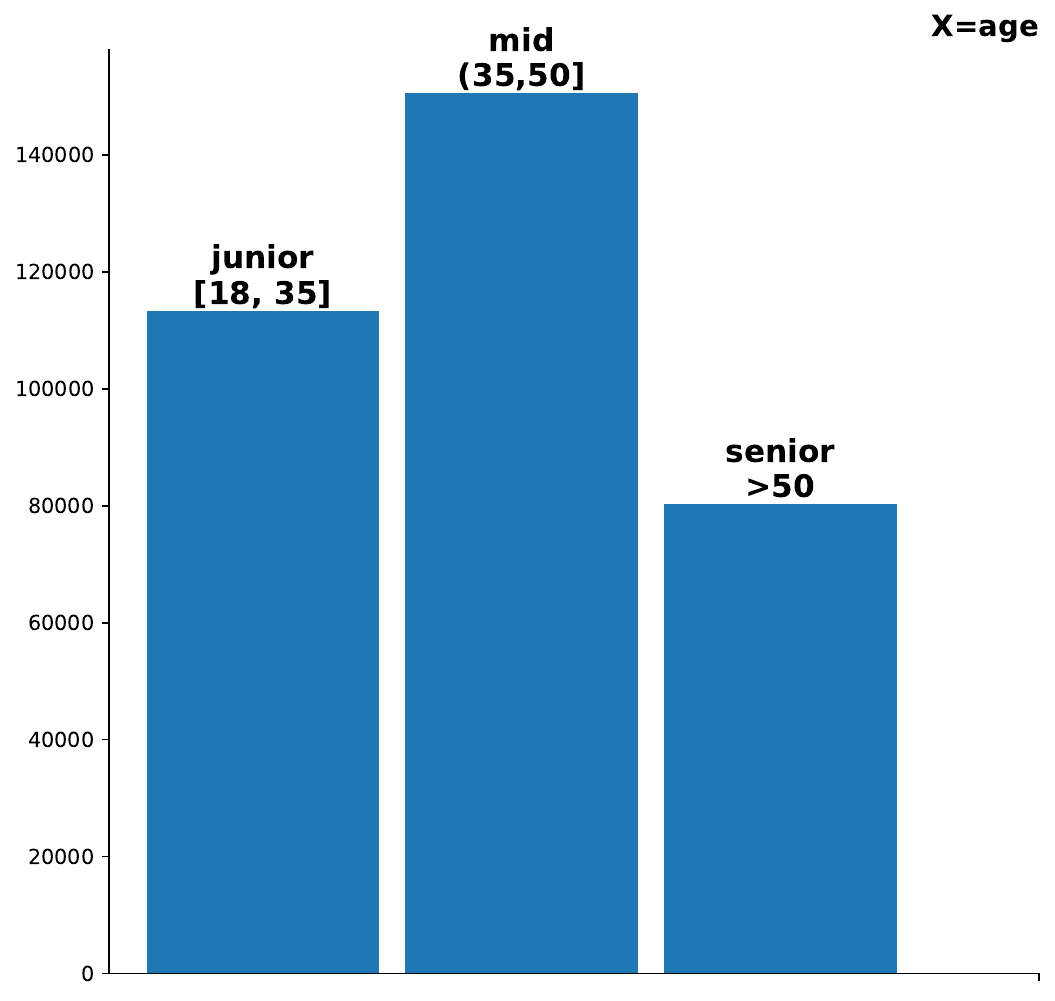}~
    \includegraphics[scale=0.3]{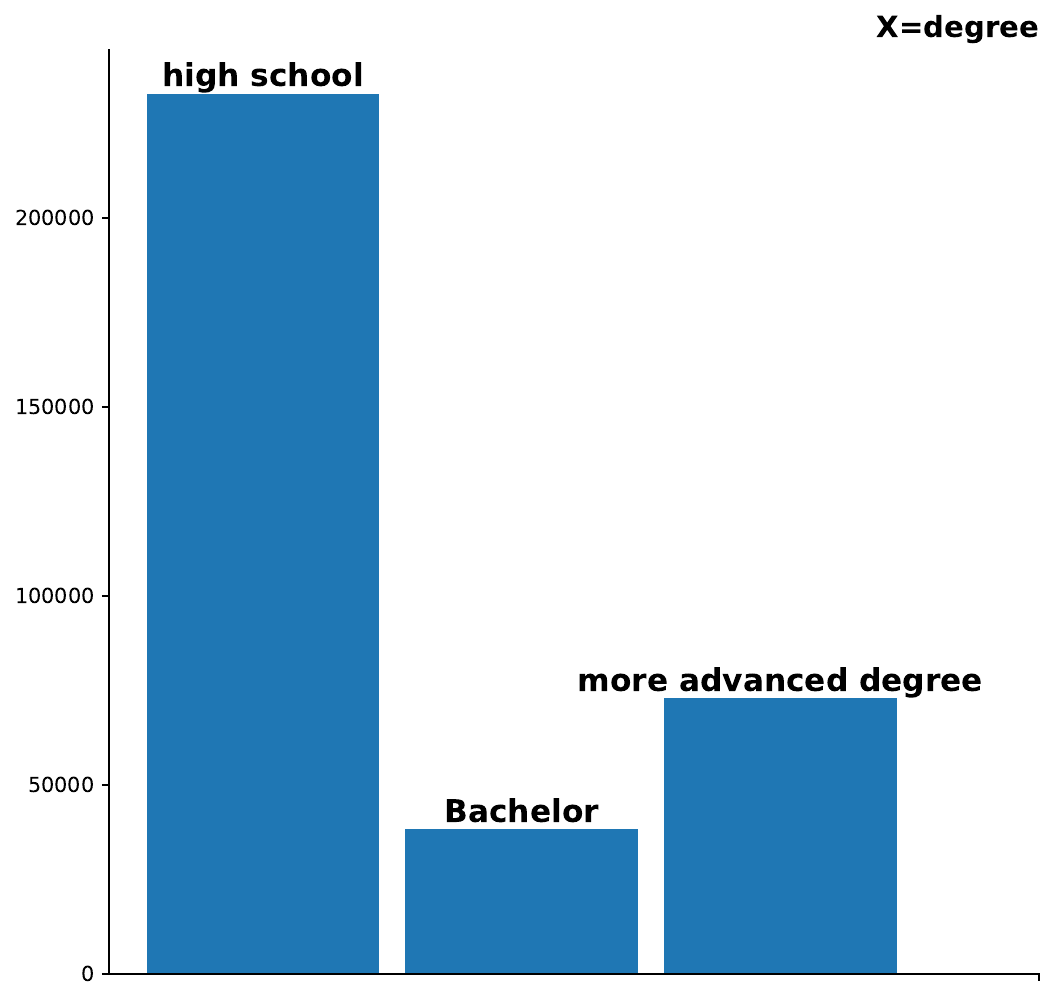}\\[1.5ex]
    \includegraphics[scale=0.35]{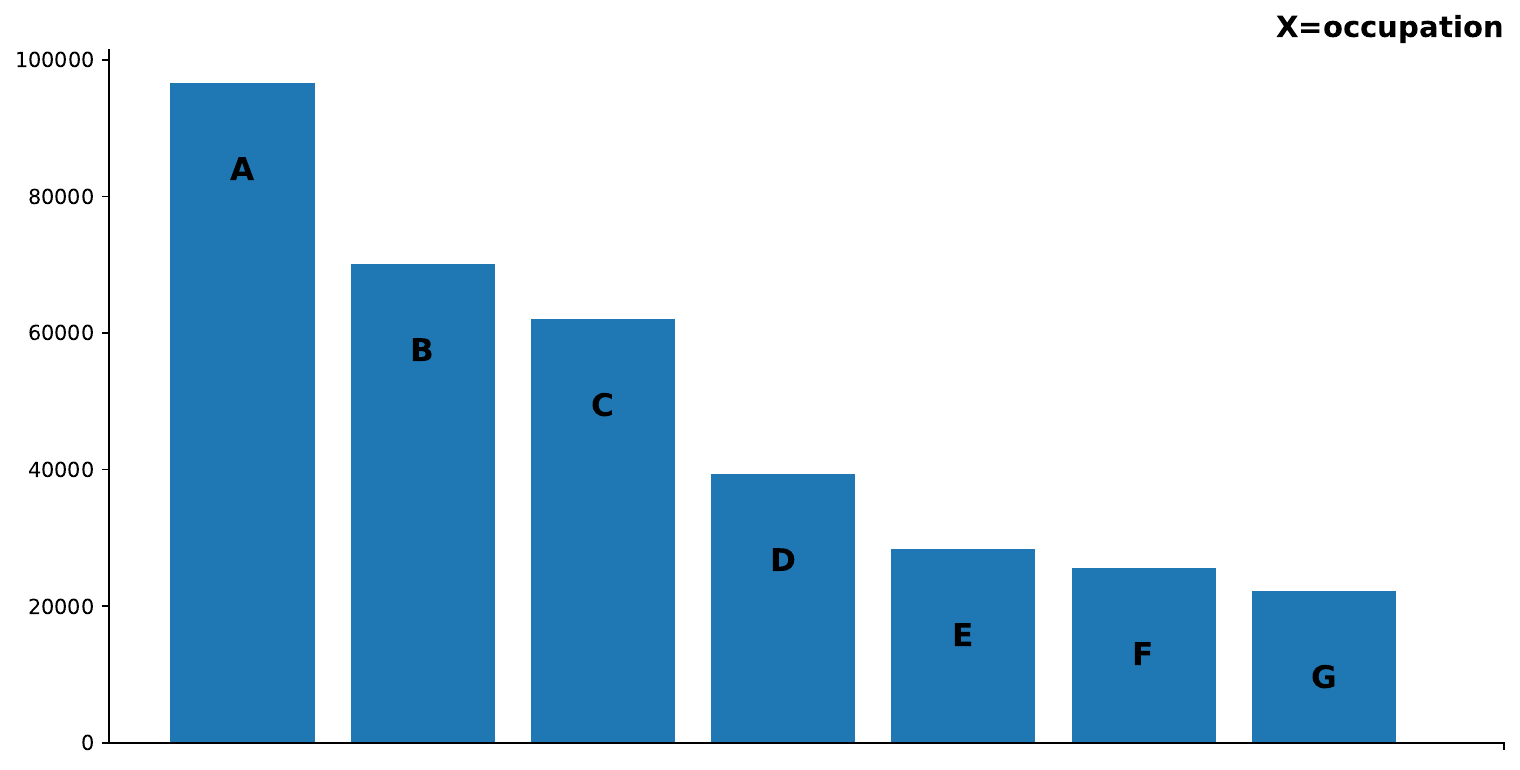}
    \caption{Cumulative prevalence (upper left) of \texttt{age\,\,groups} over all years. 
    Cumulative prevalence (upper right) of education level. 
    Cumulative prevalence (below) of A-G sectors 
    for \texttt{sales\,\&\,trade, finance\,\&\,professional\,\&\,hotels/restaurants,
       education\,\&\,social/art/other, medical,
       transport\,\&\,communications\,\&\,utilities,
       agriculture\,\&\,mining/construction, public administration}.}
    \label{fig:X_prevalence}
\end{figure}
\begin{figure}[t]
    \centering
    \includegraphics[scale=0.45]{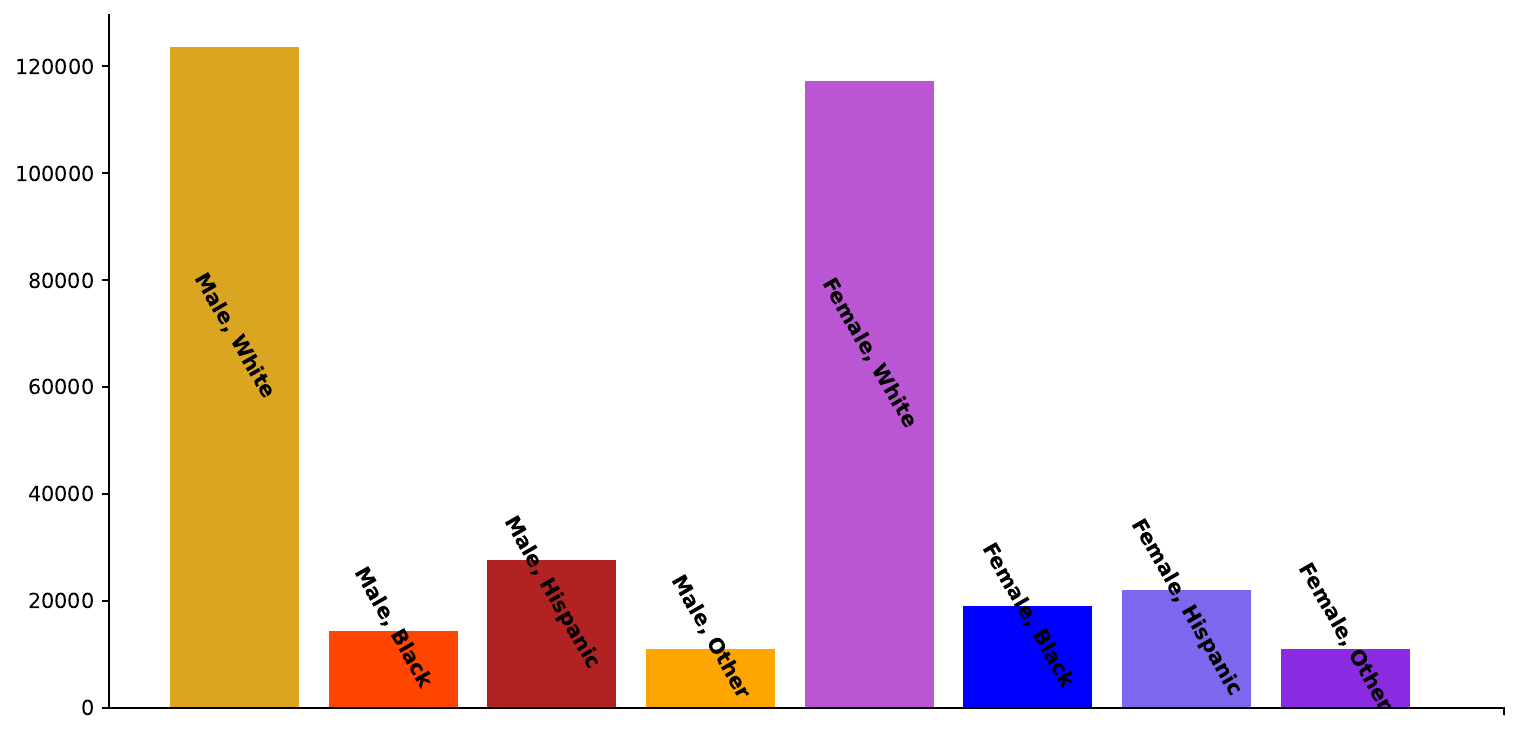}
    \caption{Cumulative prevalence of the sensitive profiles.}
    \label{fig:S_prevalence}
\end{figure}

From $N_\textrm{year} = 42\,379$, $45\,033$, $37\,144$, $56\,467$, $55\,617$, $53\,857$ and $53\,790$ census records\footnote{\url{https://www.kaggle.com/fedesoriano/gender-pay-gap-dataset}}
for the years $1981$, $1990$, $1999$, $2007$, $2009$, $2011$ and $2013$, respectively, it is straight-forward to define categorical predictor attributes. 
The cumulative statistics  of unprotected  predictors over all years are presented in bar plot~\ref{fig:X_prevalence} alongside the prevalence of protected attributes $\mathbf S = \texttt{gender, ethnicity}$ in~\ref{fig:S_prevalence}.
As response variable $Y$, we use the hourly wage adjusted for 2010 inflation.  

\begin{figure}[p]
    \centering
    \includegraphics[scale=0.52]{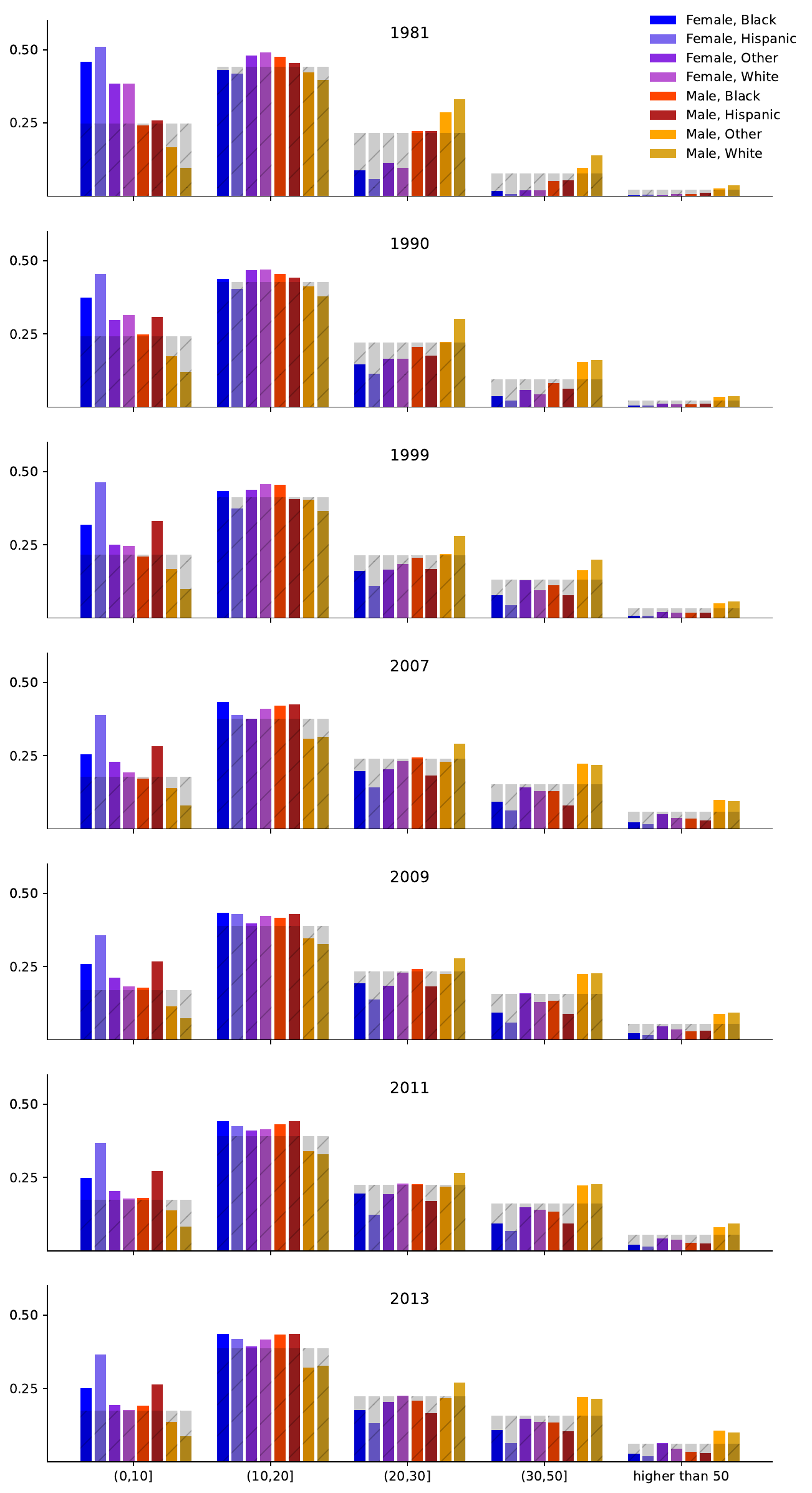}
    \caption{Conditional probability of hourly wage given sensitive profile estimated from empirical distribution $f$ (colored) and \textsc{pur} projection $q$ (gray) over the census years.}
    \label{fig:dis_parity}
\end{figure}

As a first step towards demographic fairness, we observe the disparity $f(y\vert \mathbf s)$ over the five sensible salary ranges in Figure~\ref{fig:dis_parity} for all  census years.
In principle, we could have chosen another binning for hourly wages $y$ as sensible domain $Y$ for our response variable. The mathematical guarantees of Section~\ref{sc:theory} assert the validity of \textsc{pur} methodology. 
The purpose of the provided binning is to highlight the difference in salary distribution within the lower and mid ranges, while grouping together into a larger category higher salaries that are less common in the population.

To illustrate the exact restoration of demographic Parity achieved by \textsc{pur} methods,  we present in the same Figure the fair estimate $q(y\vert \mathbf s)$ of \textsc{pur} projection.
Any dataset of size $\tilde N$ that is later sampled from $q$ is asymptotically anticipated as $\tilde N\rightarrow\infty$ to fully restore Parity in the depicted way. 
Incidentally, one could observe in Figure~\ref{fig:dis_parity} the evolution of (dis)parity over census years. \textsc{pur} methodology corrects any discriminatory biases in the protected attributes regardless of the specific circumstances described by the yearly data. 

Clearly, $q$ does not try to correct any imbalances in the distribution of salaries, which is far from uniform, given other social, educational and economical aspects.
Contrarily, the \textsc{pur} methodology uniquely unveils the joint distribution  $q$ that satisfies $q(y\vert\mathbf s)=q(y)=f(y)$, while maintaining the closest possible alignment with the original data through the acquisition~\eqref{eq:PhenoConstraints} of demographic Utility and Realism.
In terms of optimization techniques, the minimization of \textsc{kl} divergence from empirical $f$ (or from uniform $u$) under \textsc{pur} constraints exactly accomplishes the specified objectives; and nothing more~\cite{jaynes1968prior}. 

By sampling train-test data via ${mult}(N_\textrm{year} f_\textrm{year,sim}; f_\textrm{year})$ 
at the observed sample size $N_\textrm{year}$
from the yearly empirical distribution $f_\textrm{year}$, we can always deduce the \textsc{pur} projection from (mildly regularized) train data in order to use it as a natural classifier to predict on test data:
\begin{equation}
\label{eq:PredictedDistro}
    p_\textrm{pred}\config = q_\textrm{train}(y \vert \mathbf s, \mathbf x) \cdot f_\textrm{test}(\mathbf s, \mathbf x)
\end{equation}
For $f_\textrm{train}$ and $f_\textrm{test}$ we use $f_\textrm{year,sim}$.
Marginalizing distribution~\eqref{eq:PredictedDistro} over unprotected attributes $\mathbf X$, we then obtain   
the natural prediction on test data $p_\text{pred}(y\vert \mathbf s)$ given sensitive profiles. 

\begin{figure}
    \centering
    1981\includegraphics[scale=0.9, clip, trim=0 1.8cm 0 0]{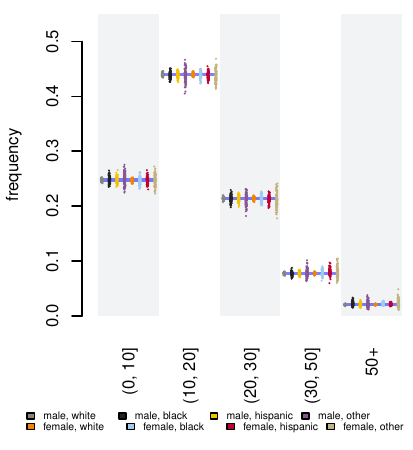}
    ~1990\includegraphics[scale=0.9, clip, trim=1.5cm 1.8cm 0 0]{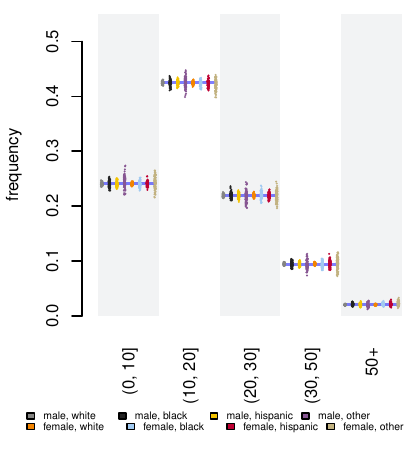}
    ~1999\includegraphics[scale=0.9, clip, trim=0 1.8cm 0 0]{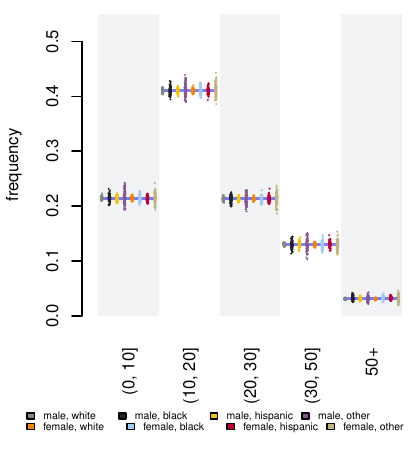}
    ~2007\includegraphics[scale=0.9, clip, trim=1.5cm 1.8cm 0 0]{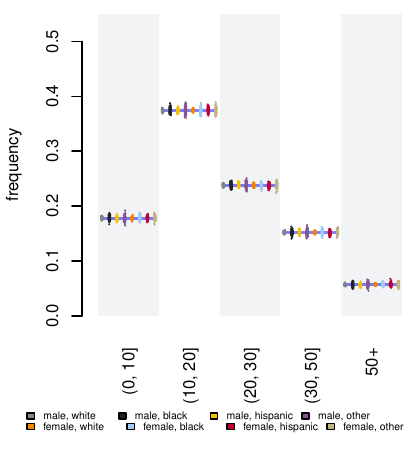}
    ~2009\includegraphics[scale=0.9, clip, trim=0 1.8cm 0 0]{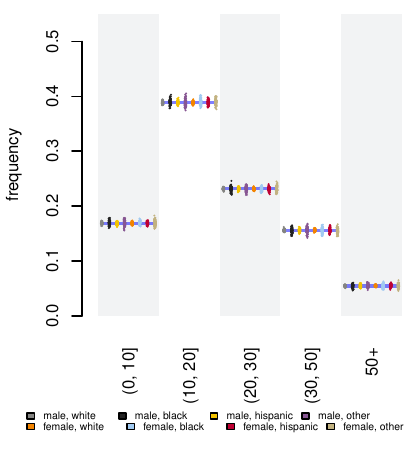}
    ~2011\includegraphics[scale=0.9, clip, trim=1.5cm 1.8cm 0 0]{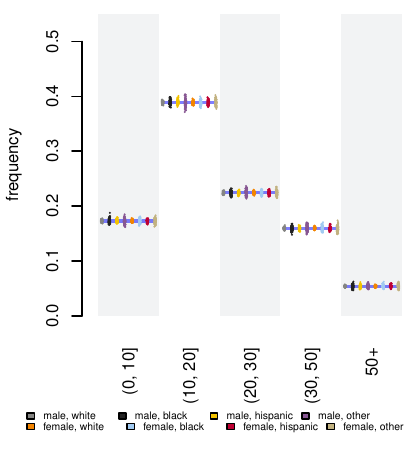}
    ~2013\includegraphics[scale=0.9]{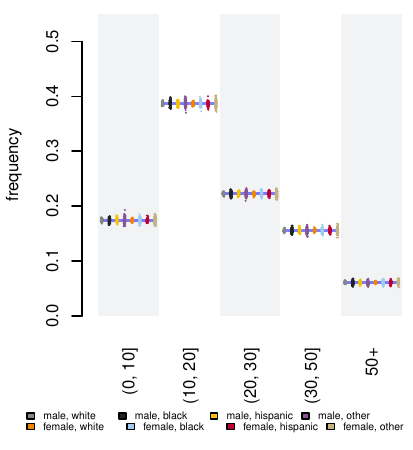}
    \caption{Natural prediction~\eqref{eq:PredictedDistro} on test data by the \textsc{pur} projection of train data in each census year.}
    \label{fig:fair-prediction}
\end{figure}

Starting from original $f_\textrm{year}$ for each census year available in the repository, Figure~\ref{fig:fair-prediction} presents the natural prediction on test data of the \textsc{pur} projection trained on 1\,000  datasets of size $N_\textrm{year}$ that were  simulated from $f_\textrm{year}$.
As expected, the prediction for hourly salary $y$ given the various social profiles $\mathbf s$ fluctuates around $f_\textrm{year}(y)$ (blue horizontal line) of the original census data. This verifies that on average the generalization of \textsc{pur} methodology is discriminatory-free regarding $\mathbf S$.
In the early census years, when greater disparities combined with a lower prevalence of higher salary ranges were encountered, the estimated values on the simulated datasets exhibit wider fluctuations.

To better comprehend the logic dictating all phenomenological constraints in~\eqref{eq:PhenoConstraints}, we plot in Figures~\ref{fig:attributable_disparity_a},~\ref{fig:attributable_disparity_b} the attributable disparity w.r.t.\ $\mathbf s_0=\texttt{male,\,\,white}$
\begin{equation*}
    p_\textrm{pred}(y\vert \mathbf s) - p_\textrm{pred}(y\vert \mathbf s_0) 
\end{equation*}
using in Eq.~\eqref{eq:PredictedDistro} different information projections of the empirical distributions $f_\textrm{train}$ describing simulated train data. 
From left to right, we learn all frequencies from $f_\textrm{train}$, hence the information projection trivially coincides with $f_\textrm{train}$ itself. Next, we only require demographic Parity  (\textsc{p}) when minimizing the $\textsc{kl}$ divergence from $f_\textrm{train}$. In the third approach, we impose demographic Parity and Utility (\textsc{pu}). Finally, we give the  attributable disparity predicted by the \textsc{pur} projection corresponding to Figure~\ref{fig:fair-prediction}.
\begin{figure}
    \centering
    1981\includegraphics[scale=0.8, clip, trim=0 1.8cm 0 0]{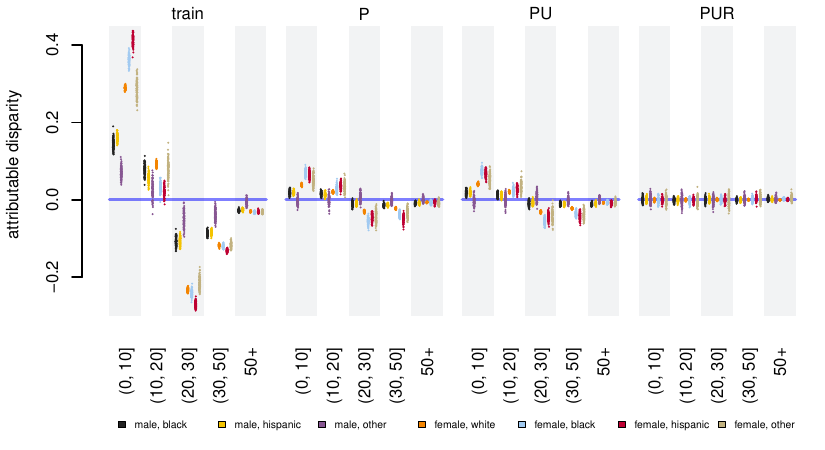}
    1990\includegraphics[scale=0.8, clip, trim=0 1.8cm 0 0]{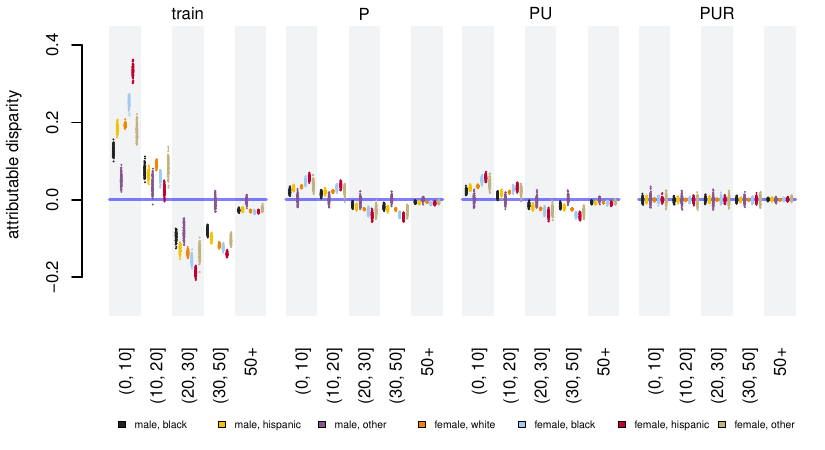}
    1999\includegraphics[scale=0.8, clip, trim=0 1.8cm 0 0]{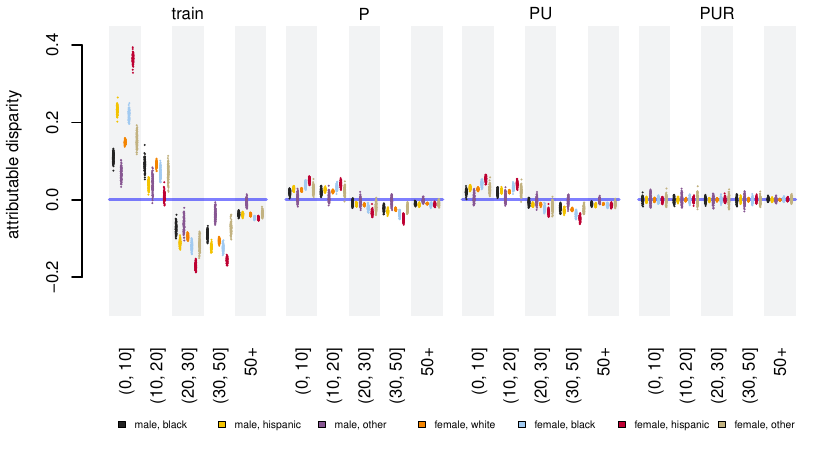}
    \caption{Attributable disparity  on simulated test data predicted by simulated train data as well as by \textsc{p}, \textsc{pu} and \textsc{pur} projections of the train data. }
    \label{fig:attributable_disparity_a}
\end{figure}
\begin{figure}
    \centering
    2007\includegraphics[scale=0.8, clip, trim=0 1.8cm 0 0]{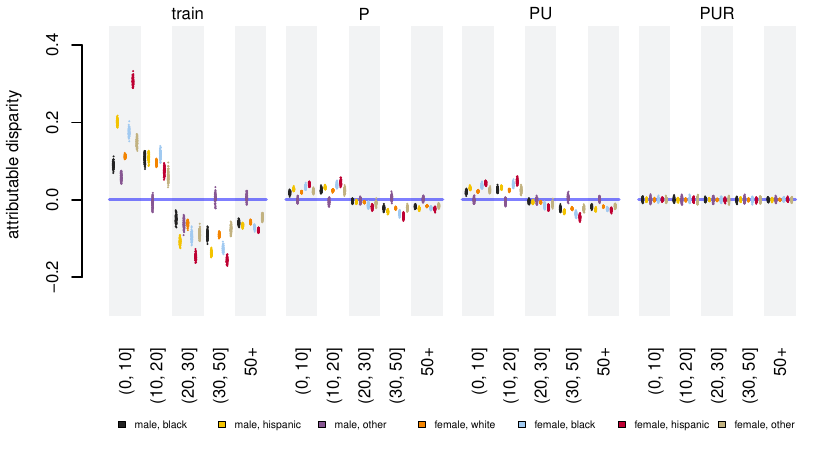}
    2009\includegraphics[scale=0.8, clip, trim=0 1.8cm 0 0]{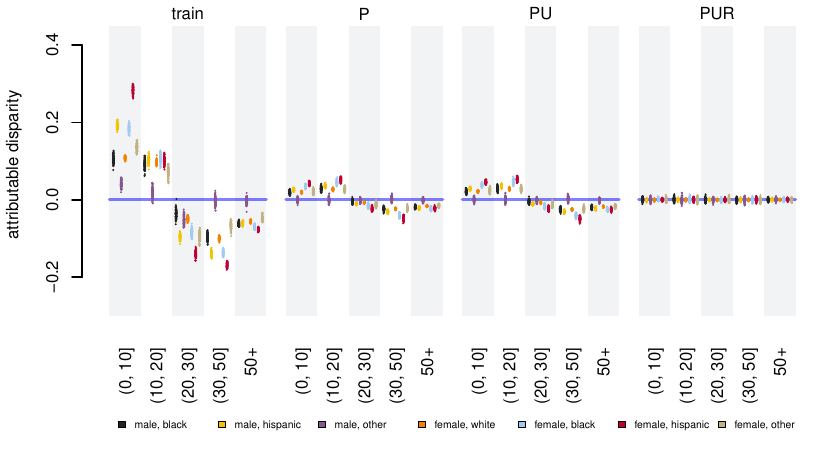}
    2011\includegraphics[scale=0.8, clip, trim=0 1.8cm 0 0]{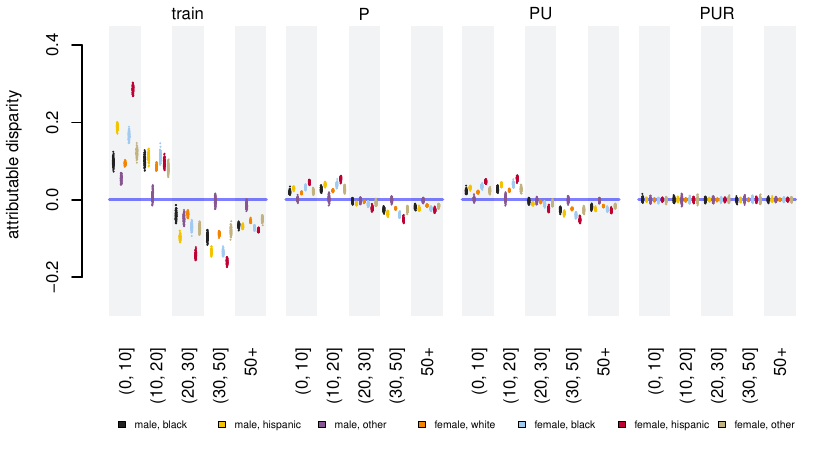}
    2013\includegraphics[scale=0.8]{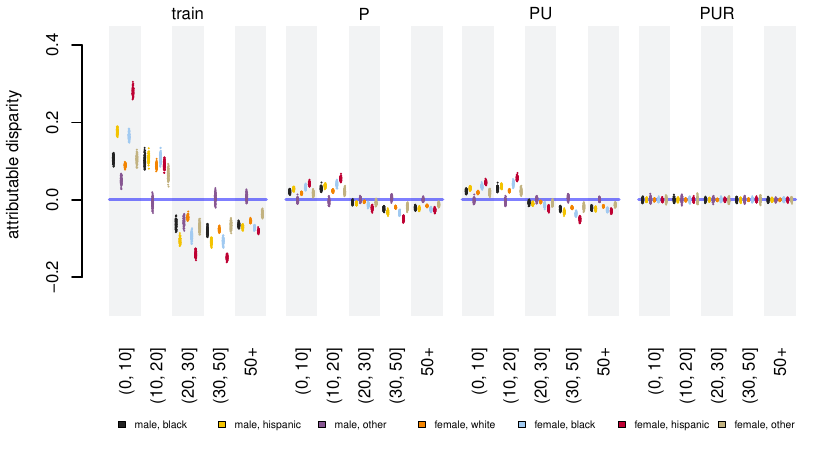}
    \caption{Continuation of~\ref{fig:attributable_disparity_a}}
    \label{fig:attributable_disparity_b}
\end{figure}
Most crucially, \textsc{pur} guarantees that up to minimal  fluctuations due to generalizing on test data whose  $f_\textrm{test}(\textbf s, \textbf x)$ is expected to slightly differ from $f_\textrm{train}(\textbf s, \textbf x)$, disparity is not re-introduced after de-biasing the train data.  
As sample sizes $N_\textrm{train}$ and $N_\textrm{test}$ grow, all fluctuations get suppressed eventually preserving demographic Parity.

\begin{figure}
    \centering
    1981\includegraphics[scale=0.9, clip, trim=0 1.8cm 0 0]{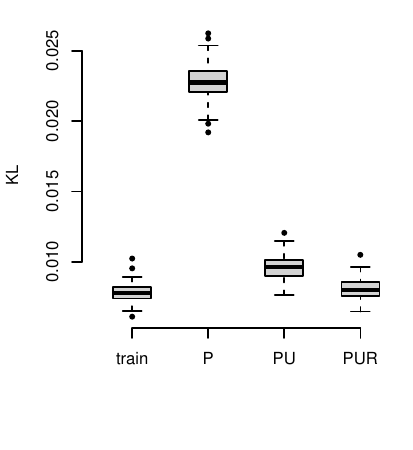}
    ~1990\includegraphics[scale=0.9, clip, trim=1.5cm 1.8cm 0 0]{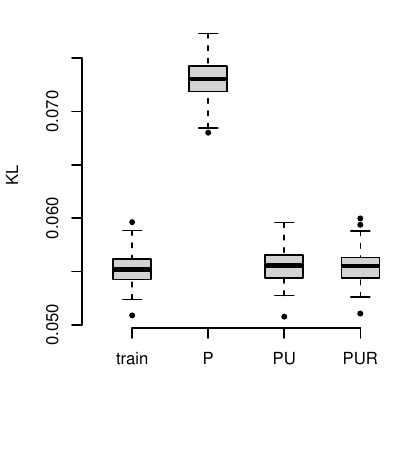}
    ~1999\includegraphics[scale=0.9, clip, trim=0 1.8cm 0 0]{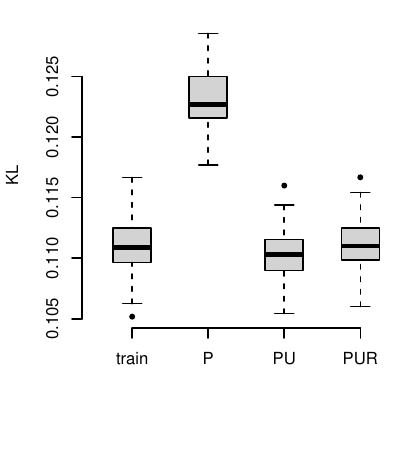}
    ~2007\includegraphics[scale=0.9, clip, trim=1.5cm 1.8cm 0 0]{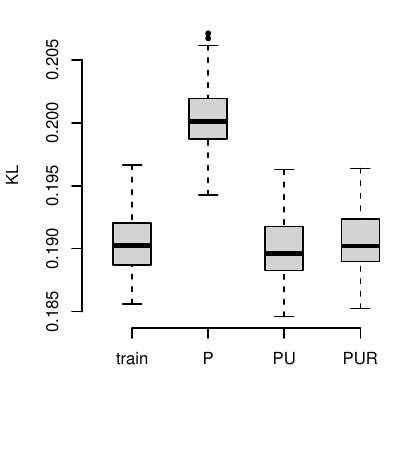}
    ~2009\includegraphics[scale=0.9, clip, trim=0 1.8cm 0 0]{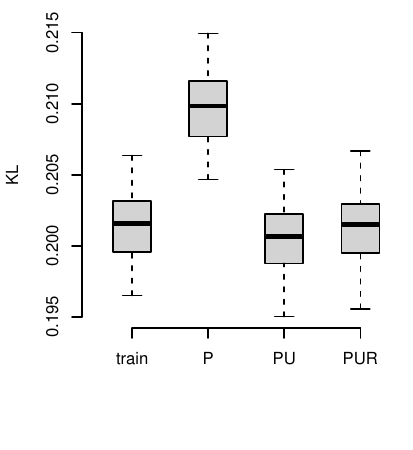}
    ~2011\includegraphics[scale=0.9, clip, trim=1.5cm 1.8cm 0 0]{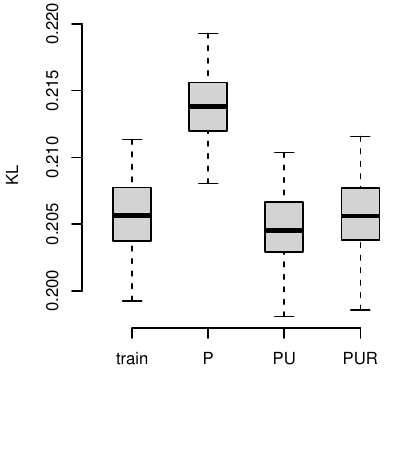}
    ~2013\includegraphics[scale=0.9, clip, trim=1.5cm 1.cm 0 0]{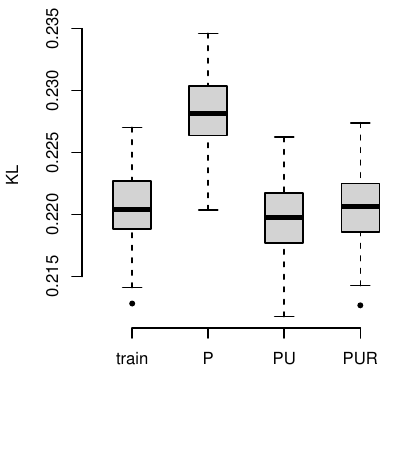}
    \caption{Box plots of the \textsc{kl} divergence of predicted from test Utility-related $Y-\mathbf X$ marginal.}
    \label{fig:utility-error}
\end{figure}

To further assess generalization capabilities of the suggested de-biasing of train data, 
Figure~\ref{fig:utility-error} lists box-plots in each census year for the Utility error (\textsc{kl} divergence of Utility marginals)
\begin{equation*}
    \sum_{y\in Y}\sum_{\mathbf x\in\mathbf X} f_\texttt{test}(y, \mathbf x) \log \frac{f_\textrm{test}(y, \mathbf x)}{p_\textrm{pred} (y, \mathbf x) }    
\end{equation*}
of the prediction~\eqref{eq:PredictedDistro} made by the de-biased distribution. Unprotected social profiles $\mathbf x\in\mathbf X$ refer to Figure~\ref{fig:X_prevalence}.
As anticipated, methods \textsc{pu} and \textsc{pur} which learn to reproduce demographic Utility from train data,  predict the lowest Utility-based \textsc{kl} divergence from test data, accordingly.

\subsection{Adult dataset}

The adult dataset\footnote{\url{https://archive.ics.uci.edu/ml/datasets/adult}} has been extensively used as a benchmark dataset, also  in the context of fair-aware machine learning.

\begin{figure}
    \centering
    \includegraphics[scale=0.5]{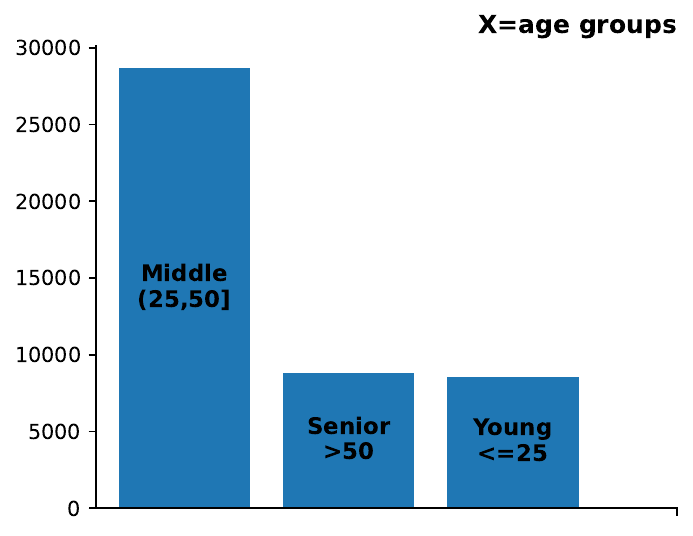}
    \includegraphics[scale=0.5]{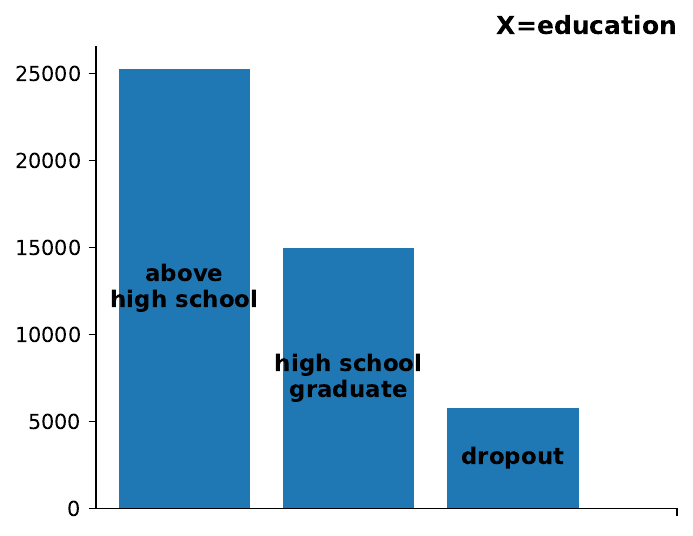}
    \includegraphics[scale=0.5]{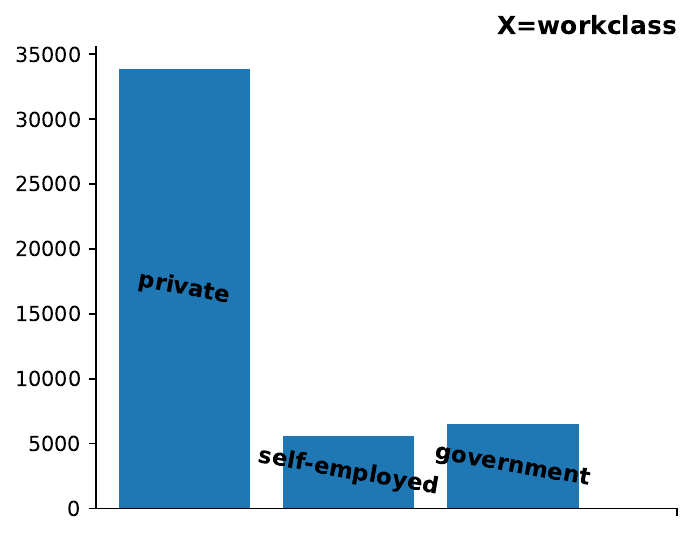}
    \includegraphics[scale=0.6]{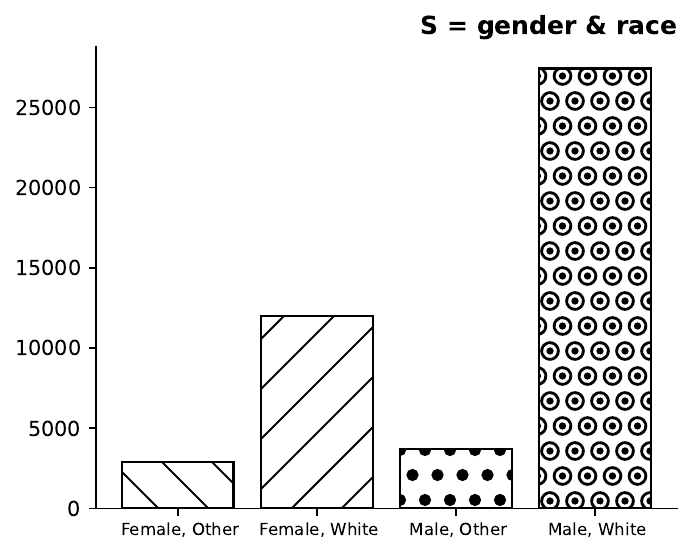}
    \caption{Prevalence of social profiles for the (un)protected predictors.}
    \label{fig:adult_dataset:predictors}
\end{figure}

\begin{figure}
    \centering
    \includegraphics[scale=0.5]{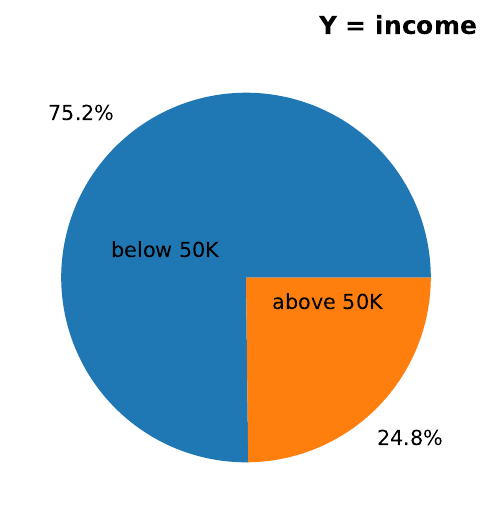}~
    \includegraphics[scale=0.7]{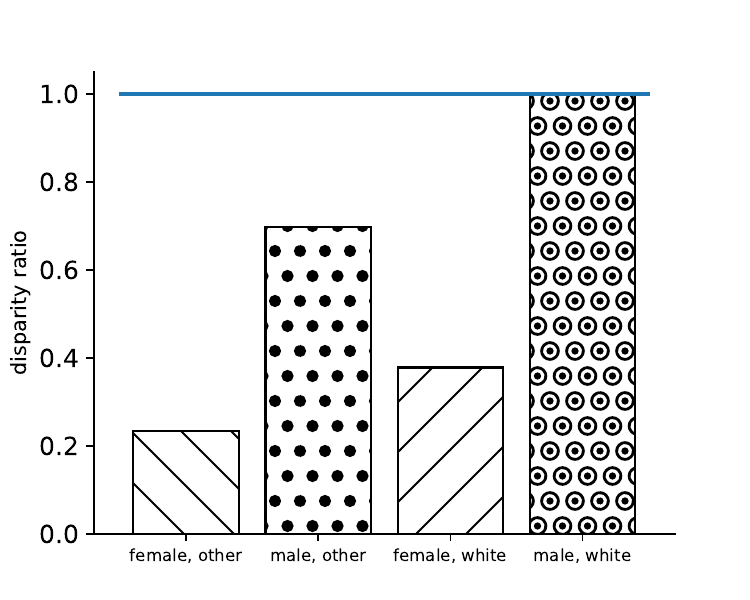}
    \caption{Prevalence (left) of income categories in the adult dataset. Disparity measure~\eqref{eq:DisparityRatio} over ratio of conditional probabilities estimated from empirical distribution.}
    \label{fig:adult-dataset:disparity}
\end{figure}

After selecting a sensible  subset of predictors $\mathbf X$ and $\mathbf S$, the statistics of the original $N_\text{data}=46\,043$ census records are summarized in Figure~\ref{fig:adult_dataset:predictors}.
Regarding the binary response $Y$, demographic disparity via the ratio
\begin{equation}
    \label{eq:DisparityRatio}
    \frac{p(y=\texttt{above 50K} \vert \mathbf s_{\phantom{0}}=\phantom{\texttt{mel}}\texttt{other}\phantom{ite})}{p(y=\texttt{above 50K} \vert \mathbf s_0=\texttt{male,\,white})}
\end{equation}
becomes clearly recognizable in Figure~\ref{fig:adult-dataset:disparity} when $p=f_\text{data}$.
Similar to multi-label classification, we aim at bringing this measure to unity for all sensitive profiles $\mathbf s$.

To serve this goal, we split the original dataset into train and test data. 
Subsequently, we minimize using the machinery of Section~\ref{ssc:IPF} the \textsc{kl} divergence from $f_\text{train}$ under demographic Parity (\textsc{p}) only as well as all $\textsc{pur}$ constraints~\eqref{eq:PhenoConstraints}.
In addition,  we minimize the \textsc{kl} divergence from the uniform distribution $u$ (equivalently maximize the entropy) under Eq.~\eqref{eq:PhenoConstraints}.
From the $\textsc{p}$ and \textsc{pur} projection of $f_\text{train}$ and the \textsc{pur} projection of $u$, many synthetic datasets of comparable sizes $N_\text{synthetic}\sim N_\text{data}$ can be easily generated.
As argued in the main text, such synthetic data is expected to be fair up to finite-$N_\text{data}$ fluctuations.

To demonstrate the coherence of our approach,  we conduct a --\,self-fulfilling from the perspective of theory~\ref{ssc:Iprojection}\,-- experiment. 
\begin{figure}
    \centering
    \includegraphics[page=2,scale=0.8,clip,trim=2.6cm 9cm 2.6cm 8cm]{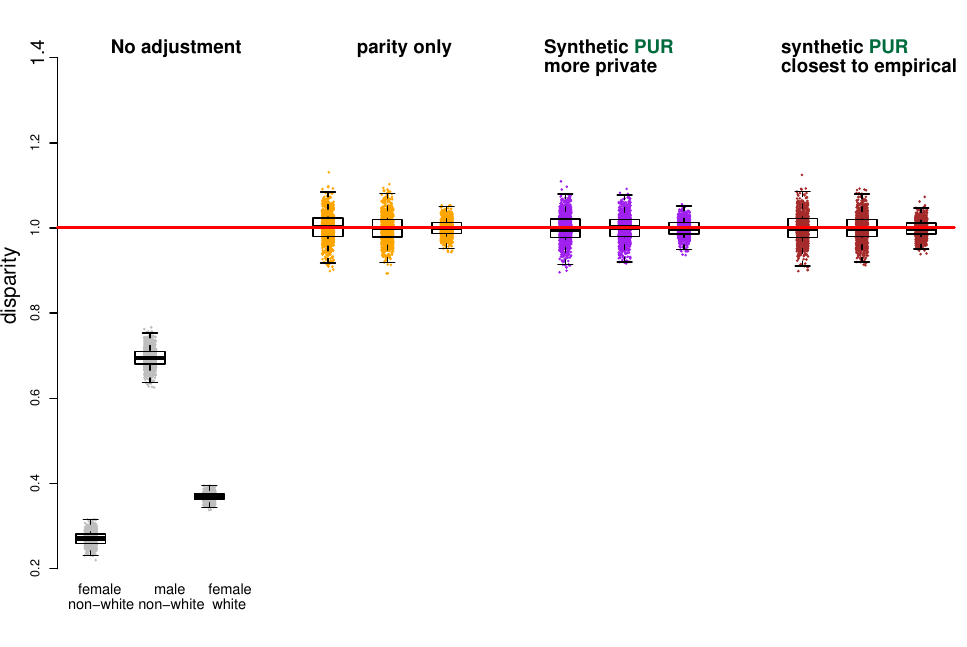}
    \caption{Disparity ratio~\eqref{eq:DisparityRatio} directly computed from the distribution of generated data.}
    \label{fig:synthetic-data}
\end{figure}
In Figure~\ref{fig:synthetic-data}, we plot the disparity~\eqref{eq:DisparityRatio} directly computed from the (re-)sampled datasets.
As a control, we utilize the disparity ratio of synthetic datasets directly generated from $f_\text{train}$ which reproduce
all the biases in the adult dataset.
On the other hand, synthetic data generated by the three methods incorporating demographic
Parity as outlined in the previous paragraph, obey on average demographic Parity. Deviations from parity attributed to
sampling noise do not fall below the 80\% threshold. 
The bias measure fluctuates in the datasets generated by de-biased distributions around 100\%, signifying that there exists no expected bias.

\begin{figure}
    \centering
    \includegraphics[page=4,scale=0.8,clip,trim=2.6cm 9cm 2.6cm 5cm]{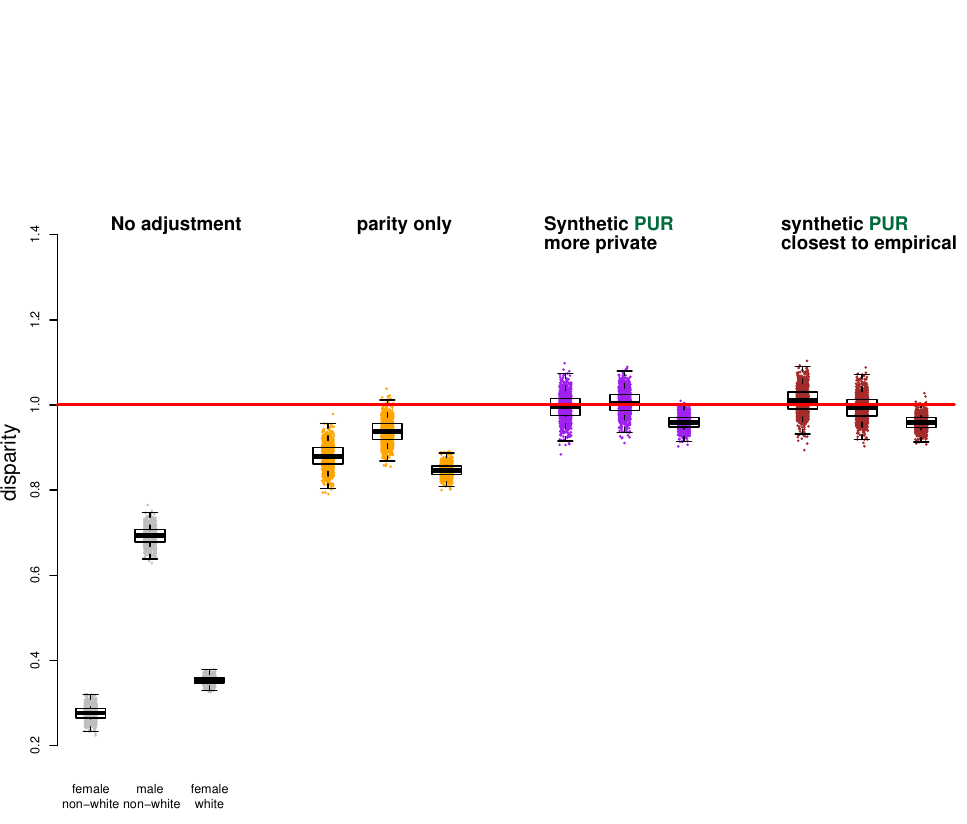}
    \caption{Disparity ratio~\eqref{eq:DisparityRatio} over sensitive profiles computed from the prediction on test data of Random Forests trained on synthetic data.}
    \label{fig:random-forests}
\end{figure}
Ultimately, we train Random Forest Classifiers (\textsc{rfc}) on our synthetic datasets in order to let them predict on $f_\text{test}(\mathbf s, \mathbf x)$.
From the outcome of the \textsc{rfc} prediction, we record in Figure~\ref{fig:random-forests} the associated disparity ratio.
To facilitate comparison of de-biasing methods, we keep over training the parameters of the classification algorithm fixed. For the purposes of fair-data generation, this is sufficient, as we do not primarily focus here on generic classification benchmarks over \textsc{ai} models.

As expected, training on re-sampled datasets from biased $f_\text{train}$ results into discriminatory \textsc{rfc}.
Since the \textsc{p} projection of $f_\text{train}$ does not incorporate information about
demographic Realism, it is not able to properly handle indirect relationships between protected attributes $\mathbf S$ and outcome $Y$ via the predictors $\mathbf X$.
Consequently, the prediction of \textsc{rfc} that has been trained on such de-biased data given test data that exhibits discriminatory relationships between the predictors re-introduces the discriminatory correlation between $\mathbf S$ and response $Y$.
Still, the disparity measure has significantly improved against training on the biased datasets. 
A similar conclusion holds for synthetic datasets generated from the $\textsc{pu}$ projection of $f_\text{train}$.

Based on the theoretical arguments of Section~\ref{sc:theory}, \textsc{rfc} trained on synthetic data
generated by methods incorporating all \textsc{pur} constraints~\eqref{eq:PhenoConstraints} remain de-biased when evaluated on $f_\text{test}(\mathbf s, \mathbf x)$, at least up to generalization errors of the implemented classifier. 
In particular, this almost optimal preservation of demographic Parity in the statistics predicted on biased test data demonstrates the merit of
incorporating demographic Realism alongside Utility during training.

\section{Code availability}
In an accompanying \texttt{Python} script, we provide auxiliary routines to compute marginal distributions, impose phenomenological constraints and run the \textsc{ipf} algorithm.
Our implementation tries to stay generic by solely relying on \texttt{numpy} and \texttt{pandas} modules.
Of course, there is room for further optimization depending on the concrete application, e.g.\ binary vs.\ multi-label classification, \textsc{pur} projection of $f$ vs.\ $u$ (\textsc{m}ax\textsc{e}nt distribution) etc. 

\bibliographystyle{abbrvnat}
{
\bibliography{PUR-2023}

}


\end{document}